\documentclass[sigconf]{acmart}
\usepackage{xcolor}
\usepackage{comment}
\usepackage{multirow}
\usepackage{subfig}
\usepackage{balance}

\AtBeginDocument{%
  \providecommand\BibTeX{{%
    \normalfont B\kern-0.5em{\scshape i\kern-0.25em b}\kern-0.8em\TeX}}}

\copyrightyear{2021}
\acmYear{2021}
\setcopyright{acmcopyright}\acmConference[MM '21]{Proceedings of the 29th ACM International Conference on Multimedia}{October 20--24, 2021}{Virtual Event, China}
\acmBooktitle{Proceedings of the 29th ACM International Conference on Multimedia (MM '21), October 20--24, 2021, Virtual Event, China}
\acmPrice{15.00}
\acmDOI{10.1145/3474085.3475354}
\acmISBN{978-1-4503-8651-7/21/10}

\begin{document}
\fancyhead{}
\title{Distributed Attention for Grounded Image Captioning}


\author{Nenglun Chen}
\affiliation{%
\institution{The University of Hong Kong}
}
\affiliation{%
\institution{Youtu Lab, Tencent}
}
% \email{chennenglun@gmail.com}

\author{Xingjia Pan}
\affiliation{%
\institution{Youtu Lab, Tencent}
}
% \email{noahpan@tencent.com}

\author{Runnan Chen}
\affiliation{%
 \institution{The University of Hong Kong}
}
% \email{rnchen2@cs.hku.hk}

\author{Lei Yang}
\affiliation{%
 \institution{The University of Hong Kong}
}
% \email{lyang@cs.hku.hk}
\author{Zhiwen Lin}
\affiliation{%
\institution{Youtu Lab, Tencent}
}
% \email{xavierzw@tencent.com}

\author{Yuqiang Ren}
\affiliation{%
 \institution{Youtu Lab, Tencent}
}
% \email{condiren@tencent.com}

\author{Haolei Yuan}
\affiliation{%
 \institution{Youtu Lab, Tencent}
}
% \email{harryyuan@tencent.com}
\author{Xiaowei Guo}
\affiliation{%
 \institution{Youtu Lab, Tencent}
}
% \email{scorpioguo@tencent.com}
\author{Feiyue Huang}
\affiliation{%
 \institution{Youtu Lab, Tencent}
}
% \email{garyhuang@tencent.com}
\author{Wenping Wang}
\affiliation{%
 \institution{The University of Hong Kong}
}
% \email{wenping@cs.hku.hk}

\renewcommand{\shortauthors}{}

\begin{abstract}
We study the problem of weakly supervised grounded image captioning. That is, given an image, the goal is to automatically generate a sentence describing the context of the image with each noun word grounded to the corresponding region in the image. This task is challenging due to the lack of explicit fine-grained region word alignments as supervision. Previous weakly supervised methods mainly explore various kinds of regularization schemes to improve attention accuracy. However, their performances are still far from the fully supervised ones. One main issue that has been ignored is that the attention for generating visually groundable words may only focus on the most discriminate parts and can not cover the whole object. To this end, we propose a simple yet effective method to alleviate the issue, termed as partial grounding problem in our paper. Specifically, we design a distributed attention mechanism to enforce the network to aggregate information from multiple spatially different regions with consistent semantics while generating the words. Therefore, the union of the focused region proposals should form a visual region that encloses the object of interest completely. Extensive experiments have demonstrated the superiority of our proposed method compared with the state-of-the-arts.
\end{abstract}



%%  \includegraphics[width=\textwidth]{sampleteaser}
%%  \caption{Seattle Mariners at Spring Training, 2010.}
%%  \Description{Enjoying the baseball game from the third-base
%%  seats. Ichiro Suzuki preparing to bat.}

\maketitle
\section{Introduction}
\label{sec:intro}

%%%%%%%%%%%%%%%%%%%%%%%%%%%%%%%%%%%%%%%%%%%%%%%%%%%%%%%%%%%%%%

Describing a given scene is considered an essential ability for humans to perceive the world. It is thus important to develop intelligent agents with such an ability, formally known as image captioning that generates the natural language descriptions based on the given images~\cite{kulkarni2013babytalk,vinyals2015show, cornia2020meshed,guo2020normalized}.
A further pursuit in the image captioning task has emerged: not only the captions should semantically reflect the content of the image, but also the words in the generated captions should be \textit{grounded} in the image, i.e., finding visual regions in the image that corresponds to the words. This way, more interpretable captions are expected to be generated.

\begin{figure}[!t]
\centering
  \includegraphics[width=0.45\textwidth]{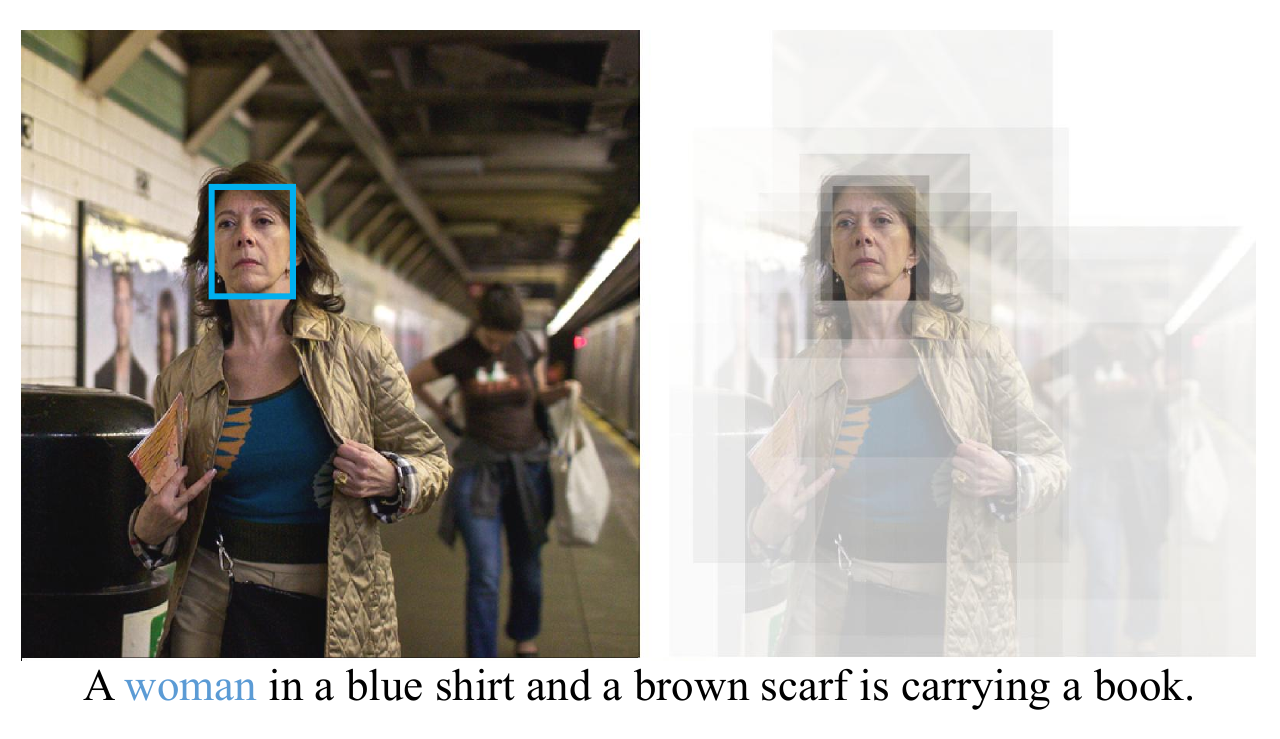}
  \caption{The partial grounding problem. The captions and the corresponding attention were generated with the baseline method. The attended region of the word 'woman' only covers the 'face'.}
  \label{fig:vis_local_grounding_att}
\end{figure}

To this end, the mainstream works leverage attention mechanisms that focus on certain regions of images to generate more grounded captions.
Pioneering works \cite{liu2017attention, yu2017supervising, zhou2019grounded} have explored using annotated bounding box as attention supervision for each visually groundable word or noun phrase in the training stage, and led to desirable improvement on the interpretability of the generated captions. However, acquiring such region-word alignment annotations is expensive and time-consuming.

Recently, several attempts have been made to generate more grounded attention with weak supervision \cite{zhou2020more, liu2020prophet, ma2020learning}. These methods mainly focus on designing various regularization schemes to produce attended regions to enclose the visual entities corresponding to the words in the generated captions. Despite the promising results achieved by the weakly supervised methods, their performances are still far from the fully supervised baselines. One main issue is that the attention for generating the visually groundable word without ground-truth grounding supervision may only focus on the most discriminant regions, failing to cover the entire object. 
See Fig. \ref{fig:vis_local_grounding_att} for example, the grounded region of the word 'woman' generated by our baseline network only covers the facial part. 
While such grounding region identify the correct person (\textit{`the woman in blue shirt'}), the network largely ignores the body parts presented in the image, which means that it fails to visually distinguish the word `woman' and the word `face'. This problem is important, especially when the input images are not accessible to the users (\textit{e.g. voice assist system for blind person}). In such cases, whether grounding on partial or the whole body may affect some follow-up decisions.

\begin{figure}
\centering
\subfloat[Statistics]{
  \includegraphics[width=0.225\textwidth]{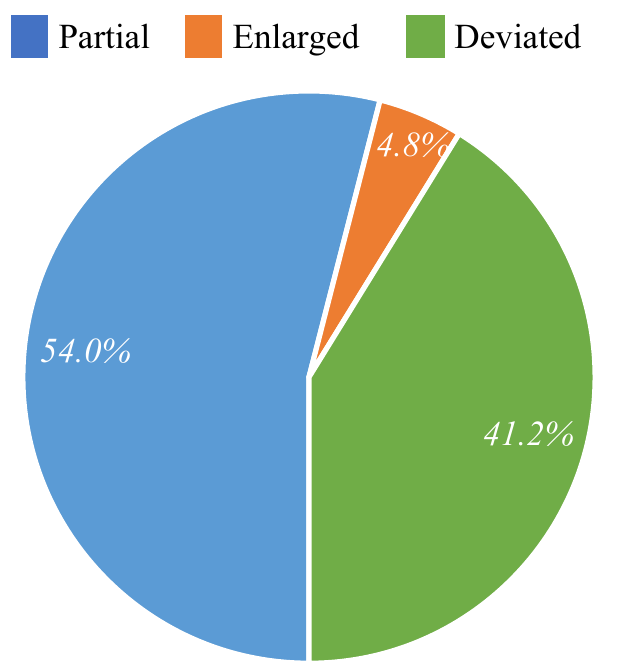}
}
\subfloat[Enlarged Grounding]{
  \includegraphics[width=0.225\textwidth]{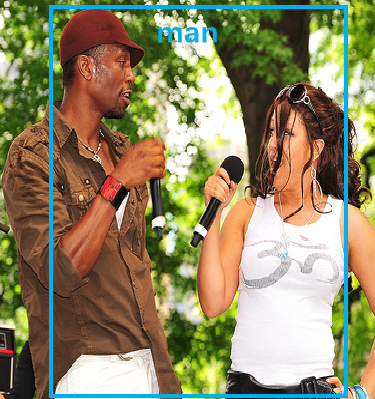}
}
\hspace{0mm}
\subfloat[Partial Grounding]{
  \includegraphics[width=0.225\textwidth]{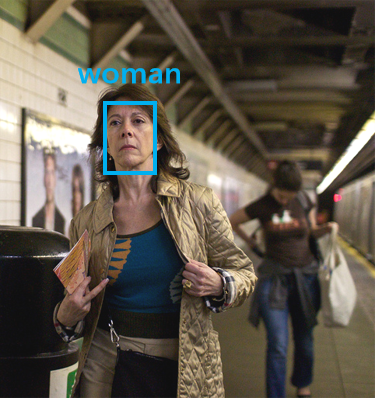}
}
\subfloat[Deviated Grounding]{
  \includegraphics[width=0.225\textwidth]{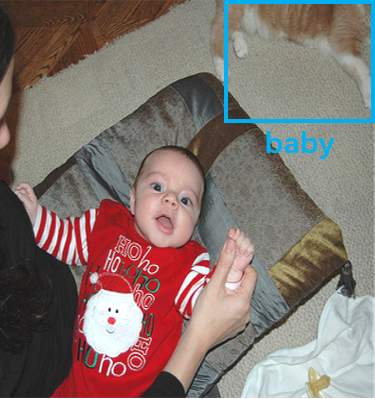}
}
\caption{Error analysis of the baseline method.}
\label{fig:vis_error_stat}
\end{figure}

Such a problem, termed as the \textit{partial grounding problem} in our paper, is usually seen in the grounded image captioning task with weak supervision. This is primarily attributed to the fact that
the objects of interest in the images are usually partially covered by many redundant region proposals, and choosing the most salient proposal inevitably leads to the described problem.
In Fig. \ref{fig:vis_error_stat}, we statistically analyze the grounding error of the baseline architecture on Flickr 30k Entities \cite{plummer2015flickr30k} dataset by casting it into three categories including partial grounding, enlarged grounding, and deviated grounding. As we can see that, the partial grounding problem contributes to a large portion of the grounding error. This further confirms the necessity to solve the partial grounding problem. 

Based on the observation, we propose a simple yet effective method to alleviate the partial grounding problem. The ingredient of our method is a distributed attention mechanism that enforces the network to attend to multiple semantically consistent regions. In this way, the union of the attended regions should form a visual region that encloses the object completely.
We follow the previous method~\cite{zhou2019grounded} and build our baseline network based on the widely used up-down architecture \cite{anderson2018bottom}. The attention module in the baseline network is augmented with multiple branches, and region proposal elimination is used to enforce different attention branches to focus on different regions.    
We conduct various experiments to verify the effectiveness of our method. 
From our experiments, we illustrate that the partial grounding problem is a crucial issue that produces inferior grounding results, and adding our distributed attention to the baseline method can effectively achieve a large performance gain on grounding accuracy than those produced by the state-of-the-art methods relying on sophisticated network designs.

In summary, our technical contributions are summarized:
\begin{itemize}
\item {We find that alleviating the partial grounding issue is critical to promote grounding performance, which is ignored in all previous methods.}

\item {We propose a novel solution to the partial grounding issue by introducing a distributed attention mechanism that enforces the network to aggregate the semantically consistent regions across the entire image.}

\item {By testing our method on the Flickr30k Entities dataset and ActivityNet Entities dataset, we achieve significant gain on grounding accuracy compared with state-of-the-art methods.}
\end{itemize} 
\section{Related Work}
\label{sec:related_work}

\paragraph{\bf Grounded Image Captioning} 
Image captioning is an important task at the juncture of computer vision and natural language processing. 
Traditional methods \cite{das2013thousand,kulkarni2013babytalk,mitchell2012midge,zhang2020relational} mainly adopts the pre-defined templates with slots that are filled with visual cues detected on the images. The performance of these methods is limited. 
With the advent of deep neural networks, the caption quality has been greatly improved. 
State-of-the-art methods \cite{yang2019auto,you2016image,lu2017knowing,anderson2018bottom,huang2019attention,guo2020normalized,cornia2020meshed} make use of various form of the attention mechanism for attending to different regions of the images while generating captions. The attention mechanism can be applied to various types of data including spatial grids \cite{lu2017knowing}, semantic metadata \cite{yang2019auto,you2016image} as well as object region proposals \cite{anderson2018bottom,huang2019attention}. Despite the great success achieved in terms of captioning performance of these methods, the grounding accuracy is still not satisfactory.
% which makes the generated captions less interpretable. 

Several attempts have been made for improving the grounding accuracy. Some pioneering works \cite{liu2017attention, yu2017supervising, zhou2019grounded,zhang2020relational} obtained improved grounding results with the help of the attention mechanism in a fully supervised manner. However, labeling the word-region correspondences is time-consuming and difficult to scale up to large datasets. Recently, researchers started to use different kinds of regularization schemes to improve the grounding accuracy in a weakly supervised manner. 
Zhou et al. \cite{zhou2020more} adopted the teacher-student scheme and distilled the knowledge learned by a Part-of-Speech enhanced image-text matching network \cite{lee2018stacked} to improve the grounding accuracy of the image captioning model. 
Ma et al. \cite{ma2020learning} employed the cyclical training regimen as a regularization, in which a generator and a localizer are trained jointly to regularize the attention. 
Liu et al. \cite{liu2020prophet} proposed the \emph{Prophet Attention}, which uses future information to compute the ideal attention and then takes the ideal attention as regularization to alleviate the deviated attention. The abovementioned methods utilized different regularization schemes to improve the grounding accuracy, however, they all neglect the partial grounding issue as described previously.
To this end, we propose a distributed attention mechanism for aggregating semantically consistent regions to alleviate the partial grounding problem.

\paragraph{\bf Visual Grounding} 
The visual grounding task aims at learning fine-grained correspondences between image regions and visually groundable noun phrases. 
Existing works can be roughly divided into two categories: supervised and weakly supervised. 
The supervised methods \cite{plummer2018conditional, liu2020learning, dogan2019neural, yu2018mattnet} took bounding boxes as supervision to enforce the alignment between image regions and noun phrases, and have achieved remarkable success. 
However, annotating the region word alignments is expensive and time-consuming, which makes it difficult to scale to large datasets. 

Weakly supervised methods \cite{liu2019knowledge,liu2021relation, rohrbach2016grounding,chen2018knowledge,gupta2020contrastive,wang2020improving} aimed at learning the correspondence with only image-caption pairs. 
Some studies utilize reconstruction regularization~\cite{rohrbach2016grounding,liu2019knowledge} to learn visual grounding. 
Rohrbach et al.~\cite{rohrbach2016grounding} first generate a candidate bounding box by the attention mechanism and then reconstruct the phrase based on the selected bounding box. 
Liu et al.~\cite{liu2019knowledge} further boost the grounding accuracy by introducing the contextual entity and model the relationship between the target entity (subject) and
contextual entity (object). 
Contrastive learning is also exploited in several method~\cite{wang2020improving,gupta2020contrastive}. 
Wang et al.~\cite{wang2020improving} learn a score function between the region-phrase pairs for distilling knowledge from a generic object detector for weakly supervised grounding. 
Gupta et al.~\cite{gupta2020contrastive} maximize a lower bound on mutual information between sets of the region features extracted from an image and contextualized word representations. 
Chen et al.~\cite{chen2018knowledge} facilitate weakly supervised grounding by considering both visual and language consistency and leveraging complementary knowledge from the feature extractor.
Our task is different from visual grounding in that our caption is to be generated rather than given for grounding.

\paragraph{\bf Weakly Supervised Object Detection} 
WSOD aims to learn the localization of objects with only image-level labels (e.g., ones given for classification) and can be roughly divided into methods based on region proposals or class activation maps (CAMs)~\cite{zhou2016learning}. Proposal-based approaches~\cite{song2014learning,gokberk2014multi,bilen2015weakly} formulate this task as multiple instance learning.
Bilen \textit{et al.}~\cite{bilen2016weakly} select proposals by parallel detection and classification branches in deep convolutional networks. 
Contextual information~\cite{kantorov2016contextlocnet}, attention mechanism~\cite{teh2016attention}, gradient map~\cite{shen2019category} and semantic segmentation~\cite{wei2018ts2c} are leveraged to learn accurate object proposals.
CAM-based methods~\cite{singh2017hide, singh2017hide, zhang2018adversarial, zhang2020inter, pan2021unveiling} produce localization maps by aggregating deep feature maps using a class-specific fully connected layer.
Despite the simplicity and effectiveness of CAM-based methods, they suffer from identifying small discriminative parts of objects. To improve the activation of CAMs, several methods~\cite{singh2017hide, yun2019cutmix, choe2019attention, zhang2018adversarial} adopted adversarial erasing on input images or feature maps to drive localization models focusing on extended object parts. 
SPG~\cite{zhang2018self} and I$^{2}$C~\cite{zhang2020inter} increased the quality of localization maps by introducing the constraint of pixel-level correlations into the network. 
SPA~\cite{pan2021unveiling} and TS-CAM~\cite{gao2021ts} obtain accurate localization maps with the help of long-range structural information. Different from the WSOD task, our task leverages the correctly attended regions for caption generation.

\section{Methodology}
\label{sec:method}

\begin{figure*}[!ht]
\centering
\subfloat[]{
  \includegraphics[width=0.66\textwidth]{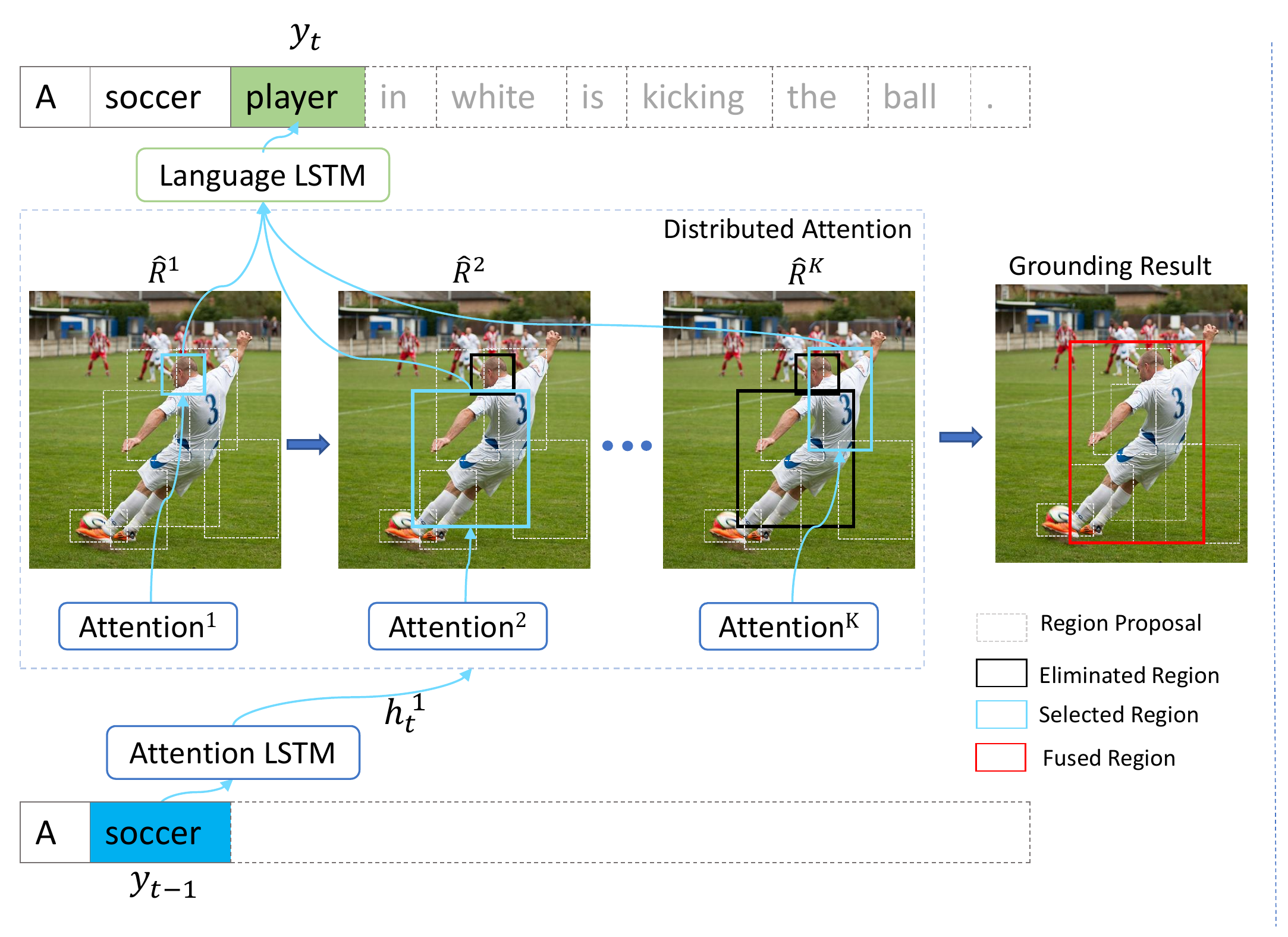}
}
\subfloat[]{
  \includegraphics[width=0.26\textwidth]{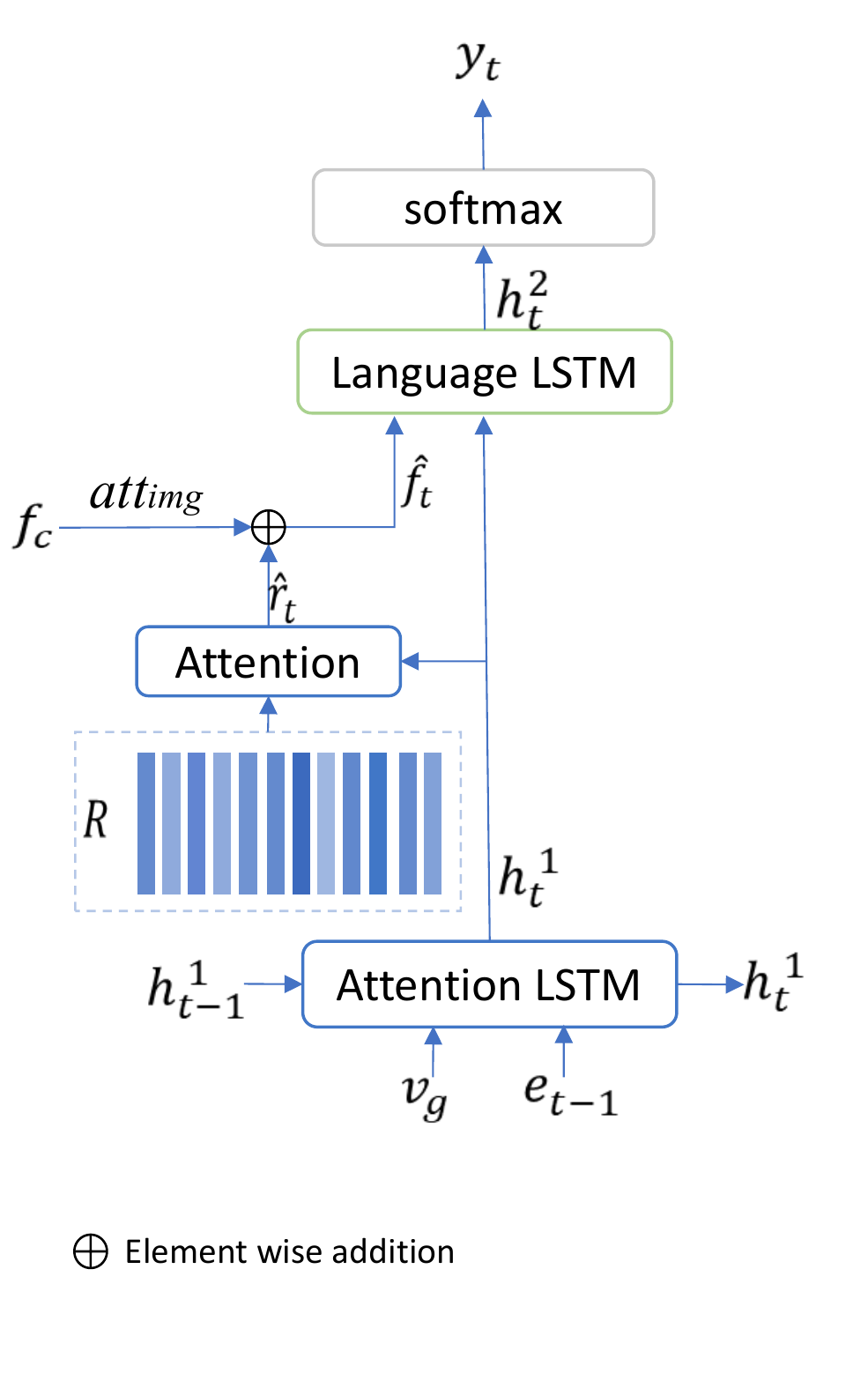}
}
  \caption{The pipeline of our proposed method (a), and the detailed framework of the baseline network (b). The attention branches in the distributed attention module are applied iteratively to discover semantically consistent regions. Note that, we do not show all the region proposals for a clear illustration.}
  \label{fig:pipeline}

\end{figure*}
%Our overall goal is to train a network to generate more grounded image captions without groundtruth grounding supervision.
In this section, we first give the notations of image captioning (Sec~\ref{sec:method_nota}).
We then briefly introduce the attention based encoder-decoder baseline network as our image captioning baseline (Sec~\ref{sec:method_review}). 
In Sec~\ref{sec:method_da}, we describe in detail the proposed distributed attention mechanism for aggregating semantically consistent partial regions for more grounded image captioning.

\subsection{Notations}
\label{sec:method_nota}
The generic image captioning task aims at generating a sentence $Y$ to describe the context of a given image $I$. We follow the pioneering works~\cite{zhou2019grounded,ma2020learning} on weakly supervised image captioning and present the input image as a set of region proposals $R=\{r_1, r_2,...,r_N\} (r_i \in \mathbb{R}^d)$ and the image feature map $f_c$, where $N$ and $d$ are the number and the feature dimension of the region proposals, respectively. We generate $r_i$ with Faster RCNN \cite{ren2015faster} while $f_c$ is extracted from a pretrained ResNet~\cite{he2016deep} model. 
The generated sentence is represented as a sequence of one-hot vectors $Y=\{y_1, y_2,...,y_T\} (y_t \in \mathbb{R}^s)$, where $T$ is the length of the sequence and $s$ is the size of the vocabulary.

\subsection{Revisit the Baseline Method}
\label{sec:method_review}
Our method is built on GVD \cite{zhou2019grounded} except for the self-attention loss for region feature embedding.
Fig. \ref{fig:pipeline}(b) shows the core part of the GVD, named language generating module, which is an extension of the widely used Bottom-up and top-down attention network \cite{anderson2018bottom} for the image captioning task. 
Specifically, the language generating model consists of two LSTM layers, the attention LSTM layer, and the language LSTM layer. The attention LSTM is used to identify the importance of different regions in the input image for generating the next word. It takes as input the global image feature vector $v_g$, the previous word embedding vector $e_{t-1}=W_e  y_{t-1}$ and the previous hidden state $h_{t-1}^1$, and encode them into current hidden state $h_t^1$:
\begin{equation}\label{eq:lstm1}
h_t^1 = LSTM_1 ([v_g;e_{t-1}], h_{t-1}^1),
\end{equation}
where $W_e$ is the learnable word embedding matrix, $[\cdot;\cdot]$ denotes the concatenation operation and we obtain $v_g$ by applying global average pooling on the image feature map $f_c$. Note that, the description of the cell states for both LSTM layers is omitted for notational clarity. 

To obtain the attention weight $\alpha_t$ over the image regions, the hidden state $h_{t}^1$ produced by the above-mentioned attention LSTM layer is further fed into an attention module followed by a softmax layer as:
\begin{equation}\label{eq:attention_w}
\begin{aligned}
z_{i,t} &= W_a tanh(W_r r_i \oplus W_h h_{t}^1),\\
\alpha_t &= softmax(z_t),
\end{aligned}
\end{equation}
\noindent where $W_a$, $W_r$ and $W_h$ are the learnable weight matrices and $\oplus$ is the element-wise addition operation. Given the attention weight $\alpha_t$, the attended region feature $\hat{r}_t$ can be calculated as the weighted combination of the regions features:
\begin{equation}\label{eq:attention_feat}
\hat{r}_t = \alpha_t^\top R
\end{equation}
where $R \in \mathbb{R}^{N \times d}$ is a feature matrix whose column represents the feature of a region proposal. Following \cite{zhou2019grounded,ma2020learning}, we also enhance the attended region feature $\hat{r}_t$ with the image features $f_c$ to get the attended image feature $\hat{f}_t$:
\begin{equation}\label{eq:convfeat}
\begin{aligned}
\hat{f}_t = \hat{r}_t + att_{img}(f_c)
\end{aligned}
\end{equation}
where $att_{img}$ is the attention block for aggregating the image feature map $f_c$ into a feature vector. 

Finally, the attended image feature $\hat{f}_t$ is fed into the language LSTM for producing the conditional probability distribution $p(y_t|y_{1:t-1})$ of the next word $y_t$ over all possible outputs as:
\begin{equation}\label{eq:lstm2}
\begin{aligned}
h_{t}^2 = LSTM_2 ([h_t^1, \hat{f}_t], h_{t-1}^2),\\
p(y_t|y_{1:t-1}) = softmax(W_o h_{t}^2),
\end{aligned}
\end{equation}
where $y_{1:t-1}$ denotes the sequence of the previously predicted words, $p(y_t|y_{1:t-1})$ denotes the conditional probability of the word $y_t$ given the previous words, $h^2_t$ is the hidden state of the language LSTM and $W_o$ is the learnable weight.

In the training phase, the network is optimized by adopting the teacher forcing strategy given the ground truth caption sequence $Y^* = \{y_1^*, y_2^*,...,y_T^*\}$. 
The training objective is to minimize the following cross-entropy loss $L_{CE}(\theta)$:

\begin{equation}\label{eq:celoss}
\begin{aligned}
L_{CE}(\theta) = -\sum_{t=1}^{T}{log(p_{\theta}(y^*_t|y^*_{1:t-1}))}
\end{aligned}
\end{equation}
where $\theta$ denotes the trainable parameters. The region with the maximum attention weight will be selected as the grounding result.

\begin{figure}[!tbp]
\centering
  \includegraphics[width=0.43\textwidth]{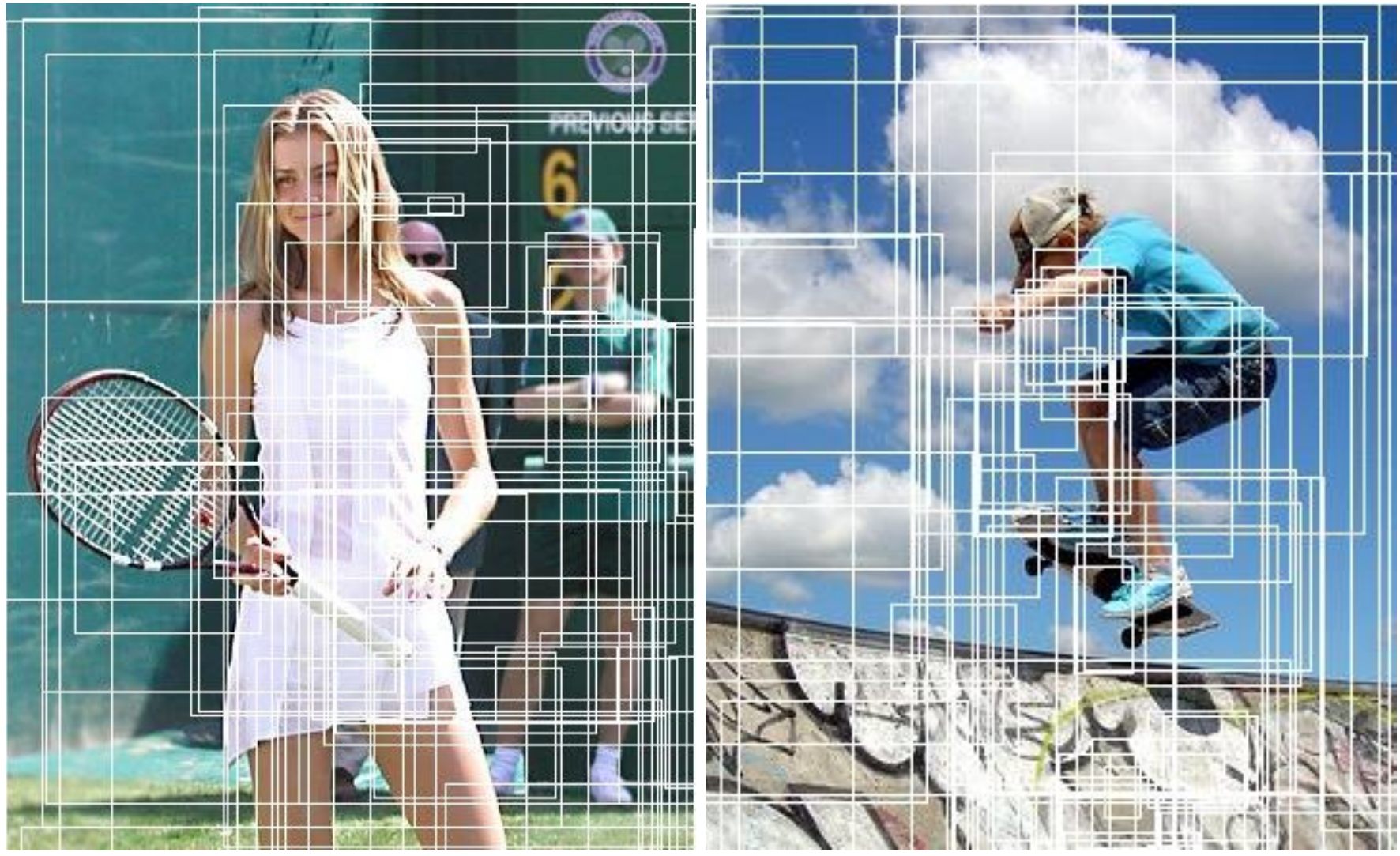}
  \caption{The extracted region proposals (white bounding boxes) are redundant and the objects are partially covered by many region proposals.}
  \label{fig:vis_bbox}
\end{figure}

\subsection{Region Aggregation with Distributed Attention}
\label{sec:method_da}

In this section, we introduce our distributed attention mechanism for aggregating semantically consistent regions to resolve the partial grounding problem in detail.

As shown in Fig. \ref{fig:vis_bbox}, the extracted region proposals by a pre-trained Faster-RCNN are redundant with each object partially covered by many region proposals. It is likely that the attention mechanism used in the baseline network will only identify parts of the objects which are the most discriminative.

Based on the above discussion, we propose the distributed attention mechanism for aggregating semantically consistent regions for generating the words. Fig. \ref{fig:pipeline}(a) illustrates the overall pipeline of our proposed method. Specifically, to make the attention focus on different regions, we augment the attention module in the baseline network with $K$ attention branches, and enforce the focused regions of all the attention branches to be mutually exclusive by iteratively applying the attention branches on the region proposals:

\begin{equation}\label{eq:mattn}
\begin{aligned}
a_t^{k} &= Attention^k(\hat{R}^k, h^1_{t-1}),\\
\hat{R}^k &= \{r_i \in R, r_i \notin M^{k-1}\}
\end{aligned}
\end{equation}
where $Attention^k()$ is the function for computing the attention weights with $k$-th attention branch over the region proposals as described in Equ. \ref{eq:attention_w}, $k \in [1,K]$ is the branch index, $\hat{R}^k$ denotes the set of region proposals used in the attention branch $k$, $a_t^k$ is the attention weights predicted by the $k$-th attention branch and $M^{k-1}$ denotes $k-1$ region proposals selected by previous $k-1$ attention branches. Note that these attention branches do not share the same weights. The attended features of all $K$ attention branches are further passed through language LSTM (Equ. \ref{eq:attention_feat}, \ref{eq:convfeat}, \ref{eq:lstm2}) to get $K$ outputs. 

At the training stage, we use the teacher-forcing strategy and apply the cross-entropy loss to all the $K$ outputs. During testing, at each time step, we select the word predicted by the most number of attention branches and combine the corresponding regions together into a single region and use it as the grounding output. 

Through our method, the distributed attention model will learn to aggregate semantically consistent regions when generating the word to alleviate the partial grounding problem. The union of the attended regions should form a visual region that encloses the object of interest completely.

\section{Experiments}
\label{sec:exp}
\subsection{Experimental Settings}
\noindent\textbf{Datasets.}
We conducted the main experiments on the widely used Flickr30k-Entities dataset \cite{plummer2015flickr30k}. It contains $31k$ images in total, and each image is annotated with $5$ sentences. In addition, it also contains $275k$ bounding box annotations and each bounding box corresponds to a visually groundable noun phrase in the caption. Following GVD \cite{zhou2019grounded}, we used the data split setting from Karpathy et al. \cite{karpathy2015deep}, which has $29k$ images for training, $1k$ images for validation, and $1k$ for testing. The vocabulary size of the dataset is $8639$.

\noindent\textbf{Evaluation Metrics.}
We used the standard captioning evaluation toolkit \cite{chen2015microsoft} to measure the captioning quality. Four commonly used language evaluation metrics are reported, i.e., BLEU \cite{papineni2002bleu}, METEOR \cite{denkowski2014meteor},
CIDEr \cite{vedantam2015cider} and SPICE \cite{anderson2016spice}.

To evaluate the grounding performance, we used the metrics $F1_{all}$ and $F1_{loc}$ defined in GVD \cite{zhou2019grounded}. For $F1_{all}$, a prediction is considered as correct if the object word is correctly generated and the Intersection-over-Union (IOU) of the predicted bounding box and the ground truth bounding box is larger than $0.5$. In $F1_{loc}$, only the correctly generated words are considered. Both metrics were averaged over classes.

\noindent\textbf{Implementation Details.}
We implemented our method with PyTorch, and all the experiments were conducted on two V100 GPUs. Following the previous works \cite{zhou2019grounded, ma2020learning, liu2020prophet}, we adopted the widely used Faster R-CNN network \cite{ren2015faster} pre-trained on Visual Genomes \cite{krishna2017visual} by GVD \cite{zhou2019grounded} to extract $100$ region proposals for each image. The word embedding dimension in the captions was set to $512$, and the word embedding layer was trained from scratch. The dimension of the hidden states for both Attention and Language LSTM was set to $1024$.

The network was optimized with Adam Optimizer \cite{kingma2014adam}, with an initial learning rate set to $5e^{-4}$ and decayed by a factor of $0.8$ for every three epochs. The batch size was set to $64$. We trained the network with a single attention branch for $20$ epoches and included the other branches in the training loop afterward. The whole training process takes less than one day. 
\begin{figure*}[ht]
\centering
  \includegraphics[width=0.98\textwidth]{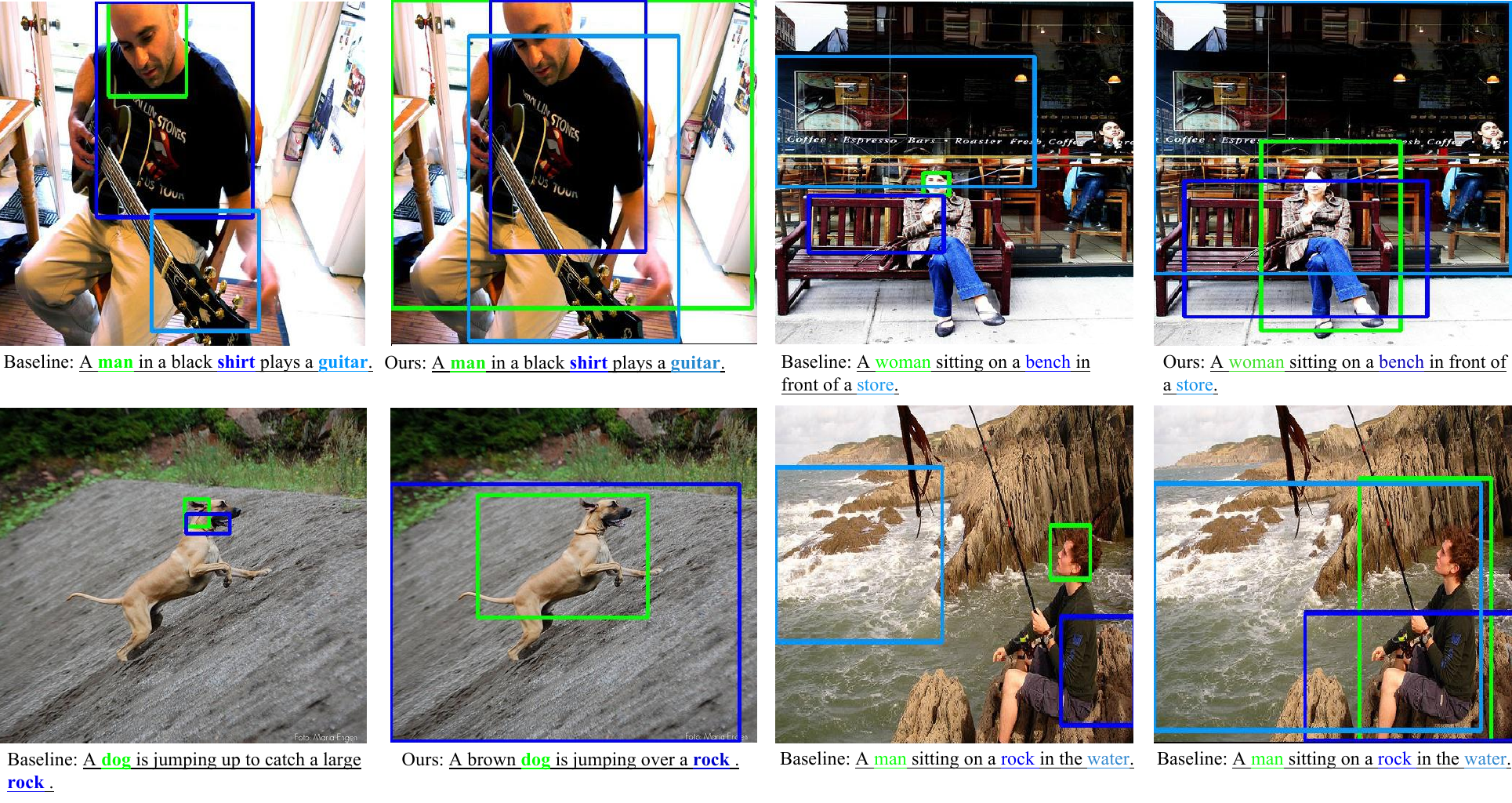}
  \caption{Results of the generated captions and the grounding regions by comparing our method with the baseline. }
  \label{fig:vis_local_grounding_cmp}
\end{figure*}

\subsection{Comparison with Prior Work}
\noindent \textbf{Quantitative Comparison.}
We compared our method with state-of-the-art weakly supervised grounded image captioning methods, including GVD \cite{zhou2019grounded}, Prophet \cite{liu2020prophet}, Zhou et al. \cite{zhou2020more} and Cyclical \cite{ma2020learning}, to see the effectiveness of our method on the grounding accuracy. 
Note that, in Zhou et al. \cite{zhou2020more}, various techniques are used to improve the captioning performance including SCST \cite{rennie2017self} and scheduled sampling \cite{bengio2015scheduled}, while we only employ the standard cross-entropy loss used in the baseline network during training. 

As shown in Table. \ref{tab:comp_sota}, by applying our proposed distributed attention module on the baseline method, we achieved significant improvement on the grounding accuracy ($F1_{all}$ and $F1_{loc}$) compared with other weakly supervised grounded image captioning methods. By comparing our method with the supervised baseline method in Cyclical, the grounding performance of our method only inferiors a little. It further confirmed the necessity of solving the partial grounding problem in this task.   

\begin{table}
  \caption{Comparison with other state-of-the-art methods on the Flickr30k Entities test set.}
  \label{tab:comp_sota}
  \resizebox{.47\textwidth}{!}{
  \begin{tabular}{lccccccc}
  \toprule
    Method&\multicolumn{5}{c}{Captioning Eval}&\multicolumn{2}{c}{Grounding Eval}\\
    \cline{2-8}
    &B@1&B@4&M&C&S&$F1_{all}$&$F1_{loc}$\\
    \midrule
    \multicolumn{2}{l}{\bf Supervised methods} \\
    Cyclical\cite{ma2020learning}&69.0&26.8&22.4&61.1&16.8&8.44&22.78\\
    GVD\cite{zhou2019grounded} &69.9&27.3&22.5&62.3&16.5&7.55&22.2\\
    \cline{1-8}
    \multicolumn{2}{l}{\bf Weakly supervised methods} \\
    GVD\cite{zhou2019grounded}&69.2&26.9&22.1&60.1&16.1&3.88&11.70\\
    Prophet\cite{liu2020prophet}&-&27.2&22.3&60.8&16.3&5.45&15.30\\
    Zhou et al.\cite{zhou2020more}&\textbf{71.4}&\textbf{28.0}&\textbf{22.6}&\textbf{66.2}&\textbf{17.0}&6.53&15.79\\
    Cyclical\cite{ma2020learning}&69.9&27.4&22.3&61.4&16.6&4.98&13.53\\
     \cline{1-8}
    Ours &69.2 & 27.2& 22.5& 62.5& 16.5& \textbf{7.91}& \textbf{21.54}\\
    \bottomrule
\end{tabular}}
\end{table}

\noindent \textbf{Qualitative Comparison.} We qualitatively compare our method with the baseline method to see how the partial grounding problem can be alleviated. As shown in Fig. \ref{fig:vis_local_grounding_cmp}, the predicted grounding regions of our method are more accurate and can cover the whole objects, while the selected regions of the baseline method may only cover the most salient parts.

\noindent \textbf{Performance on ActivityNet-Entities.}
We further tested the performance of our method on the ActivityNet-Entities dataset \cite{zhou2019grounded} to see if our method works or not on the grounded video description task. By extending GVD with our distributed attention module, we achieved significant improvement on the validation set for both $F1_{all}$(from $3.70$ to $5.32$) and $F1_{loc}$(from $12.70$ to $21.06$) with the captioning performance keeping comparable. This further confirmed the effectiveness of our proposed method.

\subsection{Qualitative Results.} 
\begin{figure*}[ht]
\centering
   \includegraphics[width=0.98\textwidth]{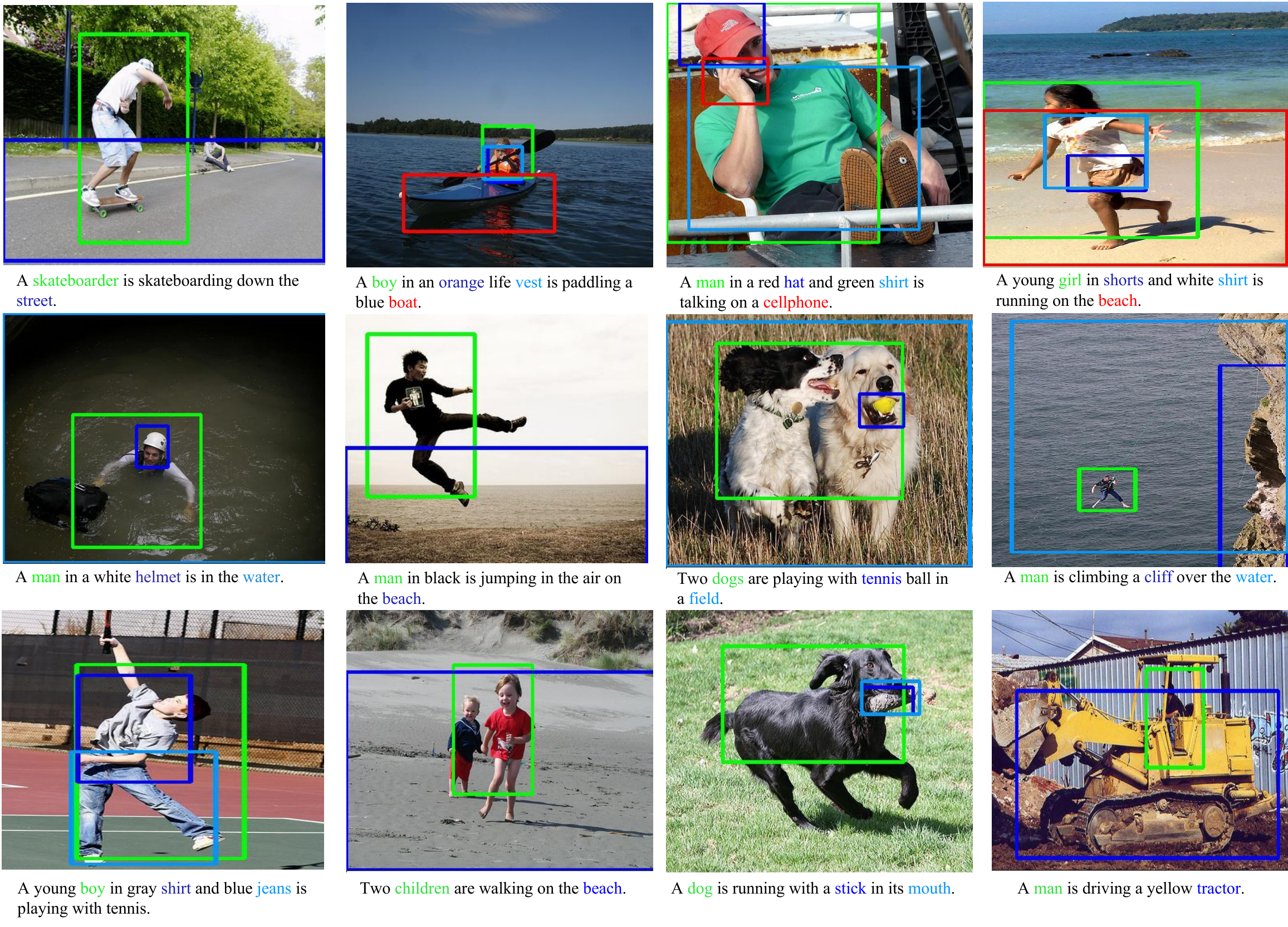}
  \caption{Visualization of our generated captions as well as the corresponding grounded regions. }
  \label{fig:vis_res}
\end{figure*}

\noindent \textbf{Example Results.} In Fig. \ref{fig:vis_res}, we present some example results including the generated captions as well as the corresponding grounding results. As we can see that, our method is able to generate fine grained image captions with accurate groundings for both foreground objects (e.g., 'man') and the background (e.g., 'beach').

\noindent \textbf{Visualization of the Distributed Attention.} We visualize the distributed attention in Fig. \ref{fig:vis_matt} to see how the region proposals are selected and fused. When generating a word, the distributed attention will attend to multiple semantically consistent regions with different locations, and the attended region proposals with the same semantic meanings will be fused to get the grounding results. We can see that our distributed attention has the ability to aggregate semantically consistent regions to enclose the object of interest completely and thus alleviate the partial grounding problem.

\subsection{Ablation Study}
\noindent\textbf{Effect of $K$.}
We tested the performance of our method with a varied number of attention branches $K$. As shown in Table \ref{tab:num_k}, the grounding performance of our method outperformed the state-of-the-art methods by a large margin when the number of attention branches $K$ is larger than $3$. When $K$ is smaller than $3$, the grounding accuracy is degraded. The main reason is that, due to the redundant region proposals extracted with Faster-RCNN, the distributed attention module is not able to discover enough regions when $K$ is too small.

\noindent\textbf{Effect of Region Proposal Elimination.}
As shown in Table \ref{tab:abl_erasing} we tested the performance of our method without region proposal elimination. Our method can also improve the grounding accuracy without region proposal elimination. That's mainly because, different initialization of the attention branches will lead to slightly different attention results, and the partial grounding problem can also be alleviated. The use of the region proposal elimination is to explicitly enforce the distributed attention module to focus on different semantically consistent regions to further boost the grounding performance.

\noindent\textbf{Relationship with Multi-head Attention.}
Our method has a close relationship with multi-head attention proposed in \cite{vaswani2017attention}. Here, we compared our distributed attention with multi-head attention by replacing the attention module in our baseline network with multi-head attention and tested it by varying the number of attention heads from $2$ to $6$. The best performing result is achieved with the number of attention head setting to $4$, that is $4.75$ for $F1_{all}$ and $13.19$ for $F1_{loc}$, which is lower than our result as shown in Table \ref{tab:comp_sota}. One possible reason is that, in our work, we explicitly enforce the attention branches to focus on multiple regions with consistent semantics, which is not guaranteed for multi-head attention.  

\begin{table}
  \caption{Effect of the number of attention branches $K$.}
  \label{tab:num_k}
  \begin{tabular}{cccccccc}
  \toprule
    K&\multicolumn{5}{c}{Captioning Eval}&\multicolumn{2}{c}{Grounding Eval}\\
    \cline{2-8}
    &B@1&B@4&M&C&S&$F1_{all}$&$F1_{loc}$\\
    \midrule
    2&68.8&26.5&22.4&60.9&16.6&5.2&15.90\\
    3&69.3&27.0&22.4&61.6&16.7&6.78&20.10\\
    4&69.2&27.2&22.5&62.5&16.5&7.91&21.54\\
    5&68.9&26.6&22.4&62.4&16.7&7.62&20.45\\
    \bottomrule
\end{tabular}
\end{table}
\begin{table}
  \caption{Results without region proposal elimination.}
  \label{tab:abl_erasing}
  \begin{tabular}{cccccccc}
  \toprule
    K&\multicolumn{5}{c}{Captioning Eval}&\multicolumn{2}{c}{Grounding Eval}\\
    \cline{2-8}
    &B@1&B@4&M&C&S&$F1_{all}$&$F1_{loc}$\\
    \midrule
    2&68.9&27.0&22.4&61.1&16.5&5.48&15.80\\
    3&68.6&27.1&22.51&61.85&16.41&5.48&16.14\\
    4&68.9&26.8&22.4&61.5&16.6&6.82&18.91\\
    5&68.9&26.8&22.3&61.4&16.3&6.27&17.36\\

    \bottomrule
 
\end{tabular}

\end{table}

\subsection{Error Analysis}
In this section, to better understand the grounding performance, we analyzed the grounding accuracy detailly by classifying all the predictions on Flickr30k Entities test set into five categories:

\begin{figure*}[!ht]
\centering
 \includegraphics[width=0.98\textwidth]{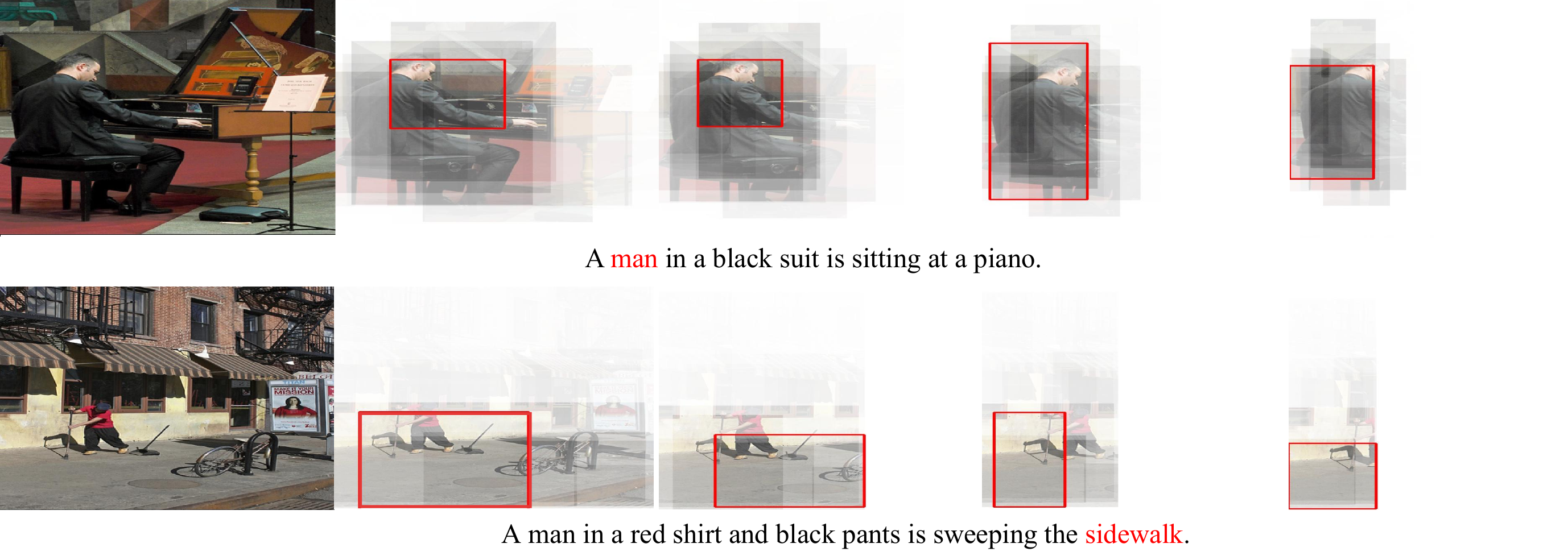}
  \caption{Visualization of the distributed attention. For a noun word (red) in a generated caption, the corresponding attention weights and the selected regions are shown. }
  \label{fig:vis_matt}

\end{figure*}
\textbf{Mis Cls}: Missing Class computes the ratio of noun words that are in the ground truth reference captions but not in the corresponding predicted captions.

\textbf{Hallu Cls}: Hallucinated Class computes the ratio of noun words that are in the predicted captions but not in the corresponding ground truth reference captions.

\textbf{Corr Grd}: Correct Grounding computes the ratio of noun words that are correctly generated and grounded ($IOU > 0.5$).

\textbf{Part Grd}: Partial Grounding computes the ratio of noun words that are correctly generated but partially grounded (the predicted bounding box is contained in the ground truth bounding boxes, but the $IOU$ of them is less than $0.5$).

\textbf{Ohter Err}: Other Error Grounding computes the ratio of noun words that are correctly generated but are neither in Part Grd nor in Corr Grd.

\begin{table}
  \caption{Detailed analysis of the grounding accuracy by classifying all the predictions on Flickr30k Entities test set into five types. The ratio of each type is computed.}
  \label{tab:ground_err_stat}
  \begin{tabular}{cccccc}
  \toprule
  &\multicolumn{2}{c}{Err Cls}&\multicolumn{3}{c}{Corr Cls}\\
    \cmidrule(r){2-3} \cmidrule(r){4-6}
    &Mis Cls&Hallu Cls&Corr Grd&Part Grd&Ohter Err\\
    \midrule
    Baseline&33.26\%&\textbf{44.92\%}&7.12\%&7.94\%&\textbf{6.76\%} \\
    Ours&\textbf{32.76\%} & 45.32\% & \textbf{10.92\%} &\textbf{2.03\%} & 8.97\% \\
    \bottomrule
\end{tabular}
\end{table}

As shown in Table \ref{tab:ground_err_stat}, the classification error (missing class and hallucinated class) contributes to the most portion of the error when computing the grounding accuracy. One main reason is that, for a given image, there exist various ways for describing its content, and directly comparing the generated captions with the corresponding references may lead to errors in computing the classification accuracy. That's why the metric $F1_{all}$, which considers the classification accuracy, is always much lower than $F1_{loc}$. And it's also the limitation of current evaluation metrics in this task.

When only considering the predictions whose classification labels are correct, our proposed method achieved much more correctly grounded predictions than that of the baseline method. It further confirmed the importance of solving the partial grounding issue and the superiority of our proposed method.  

We also measure the grounding accuracy of the baseline method if regarding all the predictions which are partially grounded as correct predictions to see the upper bound. In this case, the $F1_{all}$ and $F1_{loc}$ are $10.65$ and $30.28$ correspondingly. By comparing it with our results ($F1_{all}:7.91$, $F1_{loc}:21.5$), we can see that, there is still a large space for improvement.

\subsection{Limitations}

\begin{figure}[htbp]
\centering
  \includegraphics[width=0.45\textwidth]{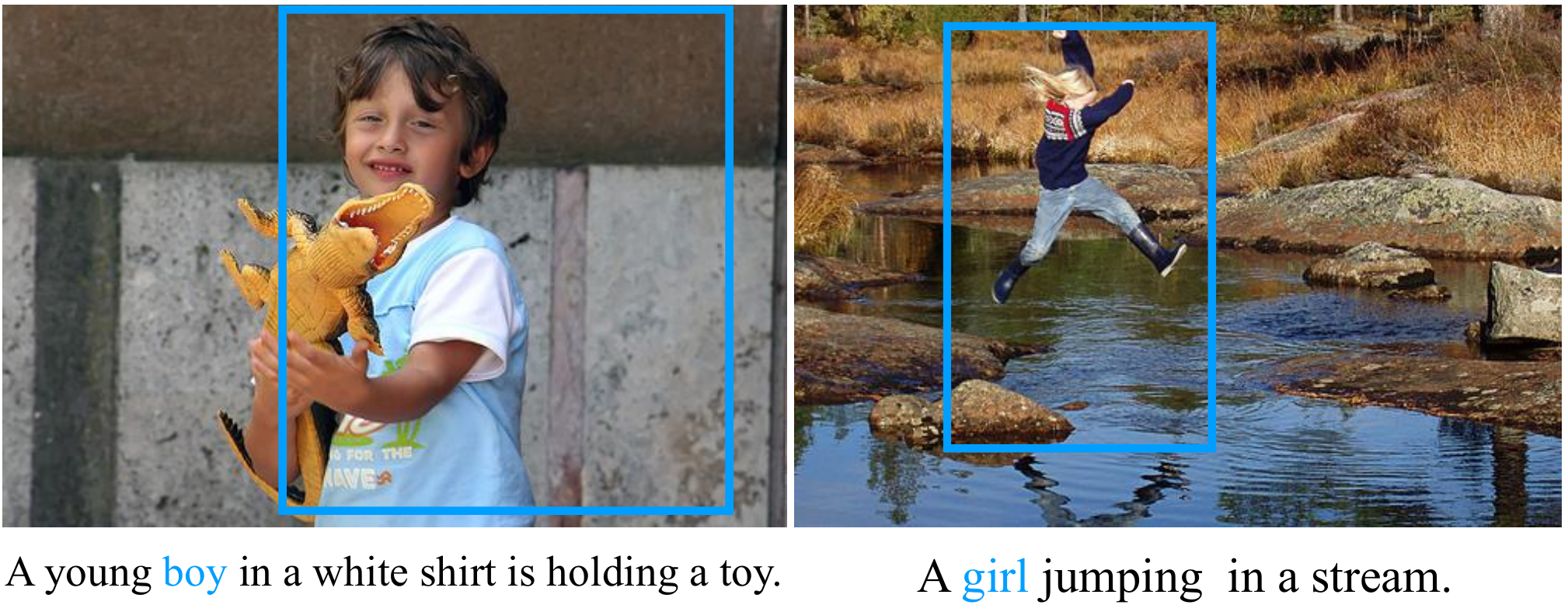}

  \caption{Limitation of our method.}
  \label{fig:vis_lim}

\end{figure}

The main limitation of our proposed method is that the predicted grounding regions might be larger than the target objects as shown in Fig. \ref{fig:vis_lim}. One important reason might be that the attention weights are calculated based on contextual information from the previous words rather than the one to be generated. Thus, the attention weights predicted by some attention branches might have the 'deviated focus' problem, and fusing the deviated regions together may lead to enlarged grounding regions. One possible solution is to combine our work with other state-of-the-arts (e.g. \cite{liu2020prophet,ma2020learning}). As it is not the main focus of our work, we leave it as future work. 

\section{Conclusion}
\label{sec:conclusion}

In this paper, we study the problem of generating image captions with accurate grounding regions without any grounding annotations. We unveil the fact that alleviating the partial grounding issue is critical to improve the grounding performance, which has been overlooked in many previous works. To this end, we propose the distributed attention mechanism to enforce the network to aggregate different regions with consistent semantics while generating the words. Extensive experiments have shown the effectiveness of our proposed method. By incorporating our proposed module into the baseline network, we achieved significant improvement on the grounding accuracy compared with the state-of-the-arts.

\bibliographystyle{ACM-Reference-Format}
\balance
\bibliography{reference}


\begin{thebibliography}{60}

%%% the Web address.

\ifx \showCODEN    \undefined \def \showCODEN     #1{\unskip}     \fi
\ifx \showDOI      \undefined \def \showDOI       #1{#1}\fi
\ifx \showISBNx    \undefined \def \showISBNx     #1{\unskip}     \fi
\ifx \showISBNxiii \undefined \def \showISBNxiii  #1{\unskip}     \fi
\ifx \showISSN     \undefined \def \showISSN      #1{\unskip}     \fi
\ifx \showLCCN     \undefined \def \showLCCN      #1{\unskip}     \fi
\ifx \shownote     \undefined \def \shownote      #1{#1}          \fi
\ifx \showarticletitle \undefined \def \showarticletitle #1{#1}   \fi
\ifx \showURL      \undefined \def \showURL       {\relax}        \fi
% invisible to TeX
\providecommand\bibfield[2]{#2}
\providecommand\bibinfo[2]{#2}
\providecommand\natexlab[1]{#1}
\providecommand\showeprint[2][]{arXiv:#2}

\bibitem[\protect\citeauthoryear{Anderson, Fernando, Johnson, and
  Gould}{Anderson et~al\mbox{.}}{2016}]%
        {anderson2016spice}
\bibfield{author}{\bibinfo{person}{Peter Anderson}, \bibinfo{person}{Basura
  Fernando}, \bibinfo{person}{Mark Johnson}, {and} \bibinfo{person}{Stephen
  Gould}.} \bibinfo{year}{2016}\natexlab{}.
\newblock \showarticletitle{Spice: Semantic propositional image caption
  evaluation}. In \bibinfo{booktitle}{\emph{European conference on computer
  vision}}. Springer, \bibinfo{pages}{382--398}.
\newblock


\bibitem[\protect\citeauthoryear{Anderson, He, Buehler, Teney, Johnson, Gould,
  and Zhang}{Anderson et~al\mbox{.}}{2018}]%
        {anderson2018bottom}
\bibfield{author}{\bibinfo{person}{Peter Anderson}, \bibinfo{person}{Xiaodong
  He}, \bibinfo{person}{Chris Buehler}, \bibinfo{person}{Damien Teney},
  \bibinfo{person}{Mark Johnson}, \bibinfo{person}{Stephen Gould}, {and}
  \bibinfo{person}{Lei Zhang}.} \bibinfo{year}{2018}\natexlab{}.
\newblock \showarticletitle{Bottom-up and top-down attention for image
  captioning and visual question answering}. In
  \bibinfo{booktitle}{\emph{Proceedings of the IEEE conference on computer
  vision and pattern recognition}}. \bibinfo{pages}{6077--6086}.
\newblock


\bibitem[\protect\citeauthoryear{Bengio, Vinyals, Jaitly, and Shazeer}{Bengio
  et~al\mbox{.}}{2015}]%
        {bengio2015scheduled}
\bibfield{author}{\bibinfo{person}{Samy Bengio}, \bibinfo{person}{Oriol
  Vinyals}, \bibinfo{person}{Navdeep Jaitly}, {and} \bibinfo{person}{Noam
  Shazeer}.} \bibinfo{year}{2015}\natexlab{}.
\newblock \showarticletitle{Scheduled sampling for sequence prediction with
  recurrent neural networks}.
\newblock \bibinfo{journal}{\emph{arXiv preprint arXiv:1506.03099}}
  (\bibinfo{year}{2015}).
\newblock


\bibitem[\protect\citeauthoryear{Bilen, Pedersoli, and Tuytelaars}{Bilen
  et~al\mbox{.}}{2015}]%
        {bilen2015weakly}
\bibfield{author}{\bibinfo{person}{Hakan Bilen}, \bibinfo{person}{Marco
  Pedersoli}, {and} \bibinfo{person}{Tinne Tuytelaars}.}
  \bibinfo{year}{2015}\natexlab{}.
\newblock \showarticletitle{Weakly supervised object detection with convex
  clustering}. In \bibinfo{booktitle}{\emph{Proceedings of the IEEE Conference
  on Computer Vision and Pattern Recognition}}. \bibinfo{pages}{1081--1089}.
\newblock


\bibitem[\protect\citeauthoryear{Bilen and Vedaldi}{Bilen and Vedaldi}{2016}]%
        {bilen2016weakly}
\bibfield{author}{\bibinfo{person}{Hakan Bilen} {and} \bibinfo{person}{Andrea
  Vedaldi}.} \bibinfo{year}{2016}\natexlab{}.
\newblock \showarticletitle{Weakly supervised deep detection networks}. In
  \bibinfo{booktitle}{\emph{Proceedings of the IEEE Conference on Computer
  Vision and Pattern Recognition}}. \bibinfo{pages}{2846--2854}.
\newblock


\bibitem[\protect\citeauthoryear{Chen, Gao, and Nevatia}{Chen
  et~al\mbox{.}}{2018}]%
        {chen2018knowledge}
\bibfield{author}{\bibinfo{person}{Kan Chen}, \bibinfo{person}{Jiyang Gao},
  {and} \bibinfo{person}{Ram Nevatia}.} \bibinfo{year}{2018}\natexlab{}.
\newblock \showarticletitle{Knowledge aided consistency for weakly supervised
  phrase grounding}. In \bibinfo{booktitle}{\emph{Proceedings of the IEEE
  Conference on Computer Vision and Pattern Recognition}}.
  \bibinfo{pages}{4042--4050}.
\newblock


\bibitem[\protect\citeauthoryear{Chen, Fang, Lin, Vedantam, Gupta, Doll{\'a}r,
  and Zitnick}{Chen et~al\mbox{.}}{2015}]%
        {chen2015microsoft}
\bibfield{author}{\bibinfo{person}{Xinlei Chen}, \bibinfo{person}{Hao Fang},
  \bibinfo{person}{Tsung-Yi Lin}, \bibinfo{person}{Ramakrishna Vedantam},
  \bibinfo{person}{Saurabh Gupta}, \bibinfo{person}{Piotr Doll{\'a}r}, {and}
  \bibinfo{person}{C~Lawrence Zitnick}.} \bibinfo{year}{2015}\natexlab{}.
\newblock \showarticletitle{Microsoft coco captions: Data collection and
  evaluation server}.
\newblock \bibinfo{journal}{\emph{arXiv preprint arXiv:1504.00325}}
  (\bibinfo{year}{2015}).
\newblock


\bibitem[\protect\citeauthoryear{Choe and Shim}{Choe and Shim}{2019}]%
        {choe2019attention}
\bibfield{author}{\bibinfo{person}{Junsuk Choe} {and} \bibinfo{person}{Hyunjung
  Shim}.} \bibinfo{year}{2019}\natexlab{}.
\newblock \showarticletitle{Attention-based dropout layer for weakly supervised
  object localization}. In \bibinfo{booktitle}{\emph{IEEE CVPR}}.
  \bibinfo{pages}{2219--2228}.
\newblock


\bibitem[\protect\citeauthoryear{Cornia, Stefanini, Baraldi, and
  Cucchiara}{Cornia et~al\mbox{.}}{2020}]%
        {cornia2020meshed}
\bibfield{author}{\bibinfo{person}{Marcella Cornia}, \bibinfo{person}{Matteo
  Stefanini}, \bibinfo{person}{Lorenzo Baraldi}, {and} \bibinfo{person}{Rita
  Cucchiara}.} \bibinfo{year}{2020}\natexlab{}.
\newblock \showarticletitle{Meshed-memory transformer for image captioning}. In
  \bibinfo{booktitle}{\emph{Proceedings of the IEEE/CVF Conference on Computer
  Vision and Pattern Recognition}}. \bibinfo{pages}{10578--10587}.
\newblock


\bibitem[\protect\citeauthoryear{Das, Xu, Doell, and Corso}{Das
  et~al\mbox{.}}{2013}]%
        {das2013thousand}
\bibfield{author}{\bibinfo{person}{Pradipto Das}, \bibinfo{person}{Chenliang
  Xu}, \bibinfo{person}{Richard~F Doell}, {and} \bibinfo{person}{Jason~J
  Corso}.} \bibinfo{year}{2013}\natexlab{}.
\newblock \showarticletitle{A thousand frames in just a few words: Lingual
  description of videos through latent topics and sparse object stitching}. In
  \bibinfo{booktitle}{\emph{Proceedings of the IEEE conference on computer
  vision and pattern recognition}}. \bibinfo{pages}{2634--2641}.
\newblock


\bibitem[\protect\citeauthoryear{Denkowski and Lavie}{Denkowski and
  Lavie}{2014}]%
        {denkowski2014meteor}
\bibfield{author}{\bibinfo{person}{Michael Denkowski} {and}
  \bibinfo{person}{Alon Lavie}.} \bibinfo{year}{2014}\natexlab{}.
\newblock \showarticletitle{Meteor universal: Language specific translation
  evaluation for any target language}. In \bibinfo{booktitle}{\emph{Proceedings
  of the ninth workshop on statistical machine translation}}.
  \bibinfo{pages}{376--380}.
\newblock


\bibitem[\protect\citeauthoryear{Dogan, Sigal, and Gross}{Dogan
  et~al\mbox{.}}{2019}]%
        {dogan2019neural}
\bibfield{author}{\bibinfo{person}{Pelin Dogan}, \bibinfo{person}{Leonid
  Sigal}, {and} \bibinfo{person}{Markus Gross}.}
  \bibinfo{year}{2019}\natexlab{}.
\newblock \showarticletitle{Neural sequential phrase grounding (seqground)}. In
  \bibinfo{booktitle}{\emph{Proceedings of the IEEE/CVF Conference on Computer
  Vision and Pattern Recognition}}. \bibinfo{pages}{4175--4184}.
\newblock


\bibitem[\protect\citeauthoryear{Gao, Wan, Pan, Peng, Tian, Han, Zhou, and
  Ye}{Gao et~al\mbox{.}}{2021}]%
        {gao2021ts}
\bibfield{author}{\bibinfo{person}{Wei Gao}, \bibinfo{person}{Fang Wan},
  \bibinfo{person}{Xingjia Pan}, \bibinfo{person}{Zhiliang Peng},
  \bibinfo{person}{Qi Tian}, \bibinfo{person}{Zhenjun Han},
  \bibinfo{person}{Bolei Zhou}, {and} \bibinfo{person}{Qixiang Ye}.}
  \bibinfo{year}{2021}\natexlab{}.
\newblock \showarticletitle{TS-CAM: Token Semantic Coupled Attention Map for
  Weakly Supervised Object Localization}.
\newblock \bibinfo{journal}{\emph{arXiv preprint arXiv:2103.14862}}
  (\bibinfo{year}{2021}).
\newblock


\bibitem[\protect\citeauthoryear{Gokberk~Cinbis, Verbeek, and
  Schmid}{Gokberk~Cinbis et~al\mbox{.}}{2014}]%
        {gokberk2014multi}
\bibfield{author}{\bibinfo{person}{Ramazan Gokberk~Cinbis},
  \bibinfo{person}{Jakob Verbeek}, {and} \bibinfo{person}{Cordelia Schmid}.}
  \bibinfo{year}{2014}\natexlab{}.
\newblock \showarticletitle{Multi-fold mil training for weakly supervised
  object localization}. In \bibinfo{booktitle}{\emph{Proceedings of the IEEE
  conference on computer vision and pattern recognition}}.
  \bibinfo{pages}{2409--2416}.
\newblock


\bibitem[\protect\citeauthoryear{Guo, Liu, Zhu, Yao, Lu, and Lu}{Guo
  et~al\mbox{.}}{2020}]%
        {guo2020normalized}
\bibfield{author}{\bibinfo{person}{Longteng Guo}, \bibinfo{person}{Jing Liu},
  \bibinfo{person}{Xinxin Zhu}, \bibinfo{person}{Peng Yao},
  \bibinfo{person}{Shichen Lu}, {and} \bibinfo{person}{Hanqing Lu}.}
  \bibinfo{year}{2020}\natexlab{}.
\newblock \showarticletitle{Normalized and Geometry-Aware Self-Attention
  Network for Image Captioning}. In \bibinfo{booktitle}{\emph{Proceedings of
  the IEEE/CVF Conference on Computer Vision and Pattern Recognition}}.
  \bibinfo{pages}{10327--10336}.
\newblock


\bibitem[\protect\citeauthoryear{Gupta, Vahdat, Chechik, Yang, Kautz, and
  Hoiem}{Gupta et~al\mbox{.}}{2020}]%
        {gupta2020contrastive}
\bibfield{author}{\bibinfo{person}{Tanmay Gupta}, \bibinfo{person}{Arash
  Vahdat}, \bibinfo{person}{Gal Chechik}, \bibinfo{person}{Xiaodong Yang},
  \bibinfo{person}{Jan Kautz}, {and} \bibinfo{person}{Derek Hoiem}.}
  \bibinfo{year}{2020}\natexlab{}.
\newblock \showarticletitle{Contrastive learning for weakly supervised phrase
  grounding}.
\newblock \bibinfo{journal}{\emph{arXiv preprint arXiv:2006.09920}}
  (\bibinfo{year}{2020}).
\newblock


\bibitem[\protect\citeauthoryear{He, Zhang, Ren, and Sun}{He
  et~al\mbox{.}}{2016}]%
        {he2016deep}
\bibfield{author}{\bibinfo{person}{Kaiming He}, \bibinfo{person}{Xiangyu
  Zhang}, \bibinfo{person}{Shaoqing Ren}, {and} \bibinfo{person}{Jian Sun}.}
  \bibinfo{year}{2016}\natexlab{}.
\newblock \showarticletitle{Deep residual learning for image recognition}. In
  \bibinfo{booktitle}{\emph{Proceedings of the IEEE conference on computer
  vision and pattern recognition}}. \bibinfo{pages}{770--778}.
\newblock


\bibitem[\protect\citeauthoryear{Huang, Wang, Chen, and Wei}{Huang
  et~al\mbox{.}}{2019}]%
        {huang2019attention}
\bibfield{author}{\bibinfo{person}{Lun Huang}, \bibinfo{person}{Wenmin Wang},
  \bibinfo{person}{Jie Chen}, {and} \bibinfo{person}{Xiao-Yong Wei}.}
  \bibinfo{year}{2019}\natexlab{}.
\newblock \showarticletitle{Attention on attention for image captioning}. In
  \bibinfo{booktitle}{\emph{Proceedings of the IEEE/CVF International
  Conference on Computer Vision}}. \bibinfo{pages}{4634--4643}.
\newblock


\bibitem[\protect\citeauthoryear{Kantorov, Oquab, Cho, and Laptev}{Kantorov
  et~al\mbox{.}}{2016}]%
        {kantorov2016contextlocnet}
\bibfield{author}{\bibinfo{person}{Vadim Kantorov}, \bibinfo{person}{Maxime
  Oquab}, \bibinfo{person}{Minsu Cho}, {and} \bibinfo{person}{Ivan Laptev}.}
  \bibinfo{year}{2016}\natexlab{}.
\newblock \showarticletitle{Contextlocnet: Context-aware deep network models
  for weakly supervised localization}. In \bibinfo{booktitle}{\emph{European
  Conference on Computer Vision}}. Springer, \bibinfo{pages}{350--365}.
\newblock


\bibitem[\protect\citeauthoryear{Karpathy and Fei-Fei}{Karpathy and
  Fei-Fei}{2015}]%
        {karpathy2015deep}
\bibfield{author}{\bibinfo{person}{Andrej Karpathy} {and} \bibinfo{person}{Li
  Fei-Fei}.} \bibinfo{year}{2015}\natexlab{}.
\newblock \showarticletitle{Deep visual-semantic alignments for generating
  image descriptions}. In \bibinfo{booktitle}{\emph{Proceedings of the IEEE
  conference on computer vision and pattern recognition}}.
  \bibinfo{pages}{3128--3137}.
\newblock


\bibitem[\protect\citeauthoryear{Kingma and Ba}{Kingma and Ba}{2014}]%
        {kingma2014adam}
\bibfield{author}{\bibinfo{person}{Diederik~P Kingma} {and}
  \bibinfo{person}{Jimmy Ba}.} \bibinfo{year}{2014}\natexlab{}.
\newblock \showarticletitle{Adam: A method for stochastic optimization}.
\newblock \bibinfo{journal}{\emph{arXiv preprint arXiv:1412.6980}}
  (\bibinfo{year}{2014}).
\newblock


\bibitem[\protect\citeauthoryear{Krishna, Zhu, Groth, Johnson, Hata, Kravitz,
  Chen, Kalantidis, Li, Shamma, et~al\mbox{.}}{Krishna et~al\mbox{.}}{2017}]%
        {krishna2017visual}
\bibfield{author}{\bibinfo{person}{Ranjay Krishna}, \bibinfo{person}{Yuke Zhu},
  \bibinfo{person}{Oliver Groth}, \bibinfo{person}{Justin Johnson},
  \bibinfo{person}{Kenji Hata}, \bibinfo{person}{Joshua Kravitz},
  \bibinfo{person}{Stephanie Chen}, \bibinfo{person}{Yannis Kalantidis},
  \bibinfo{person}{Li-Jia Li}, \bibinfo{person}{David~A Shamma},
  {et~al\mbox{.}}} \bibinfo{year}{2017}\natexlab{}.
\newblock \showarticletitle{Visual genome: Connecting language and vision using
  crowdsourced dense image annotations}.
\newblock \bibinfo{journal}{\emph{International journal of computer vision}}
  \bibinfo{volume}{123}, \bibinfo{number}{1} (\bibinfo{year}{2017}),
  \bibinfo{pages}{32--73}.
\newblock


\bibitem[\protect\citeauthoryear{Kulkarni, Premraj, Ordonez, Dhar, Li, Choi,
  Berg, and Berg}{Kulkarni et~al\mbox{.}}{2013}]%
        {kulkarni2013babytalk}
\bibfield{author}{\bibinfo{person}{Girish Kulkarni}, \bibinfo{person}{Visruth
  Premraj}, \bibinfo{person}{Vicente Ordonez}, \bibinfo{person}{Sagnik Dhar},
  \bibinfo{person}{Siming Li}, \bibinfo{person}{Yejin Choi},
  \bibinfo{person}{Alexander~C Berg}, {and} \bibinfo{person}{Tamara~L Berg}.}
  \bibinfo{year}{2013}\natexlab{}.
\newblock \showarticletitle{Babytalk: Understanding and generating simple image
  descriptions}.
\newblock \bibinfo{journal}{\emph{IEEE Transactions on Pattern Analysis and
  Machine Intelligence}} \bibinfo{volume}{35}, \bibinfo{number}{12}
  (\bibinfo{year}{2013}), \bibinfo{pages}{2891--2903}.
\newblock


\bibitem[\protect\citeauthoryear{Lee, Chen, Hua, Hu, and He}{Lee
  et~al\mbox{.}}{2018}]%
        {lee2018stacked}
\bibfield{author}{\bibinfo{person}{Kuang-Huei Lee}, \bibinfo{person}{Xi Chen},
  \bibinfo{person}{Gang Hua}, \bibinfo{person}{Houdong Hu}, {and}
  \bibinfo{person}{Xiaodong He}.} \bibinfo{year}{2018}\natexlab{}.
\newblock \showarticletitle{Stacked cross attention for image-text matching}.
  In \bibinfo{booktitle}{\emph{Proceedings of the European Conference on
  Computer Vision (ECCV)}}. \bibinfo{pages}{201--216}.
\newblock


\bibitem[\protect\citeauthoryear{Liu, Mao, Sha, and Yuille}{Liu
  et~al\mbox{.}}{2017}]%
        {liu2017attention}
\bibfield{author}{\bibinfo{person}{Chenxi Liu}, \bibinfo{person}{Junhua Mao},
  \bibinfo{person}{Fei Sha}, {and} \bibinfo{person}{Alan Yuille}.}
  \bibinfo{year}{2017}\natexlab{}.
\newblock \showarticletitle{Attention correctness in neural image captioning}.
  In \bibinfo{booktitle}{\emph{Proceedings of the AAAI Conference on Artificial
  Intelligence}}, Vol.~\bibinfo{volume}{31}.
\newblock


\bibitem[\protect\citeauthoryear{Liu, Ren, Wu, Ge, Fan, Zou, and Sun}{Liu
  et~al\mbox{.}}{2020a}]%
        {liu2020prophet}
\bibfield{author}{\bibinfo{person}{Fenglin Liu}, \bibinfo{person}{Xuancheng
  Ren}, \bibinfo{person}{Xian Wu}, \bibinfo{person}{Shen Ge},
  \bibinfo{person}{Wei Fan}, \bibinfo{person}{Yuexian Zou}, {and}
  \bibinfo{person}{Xu Sun}.} \bibinfo{year}{2020}\natexlab{a}.
\newblock \showarticletitle{Prophet Attention: Predicting Attention with Future
  Attention}.
\newblock \bibinfo{journal}{\emph{Advances in Neural Information Processing
  Systems}}  \bibinfo{volume}{33} (\bibinfo{year}{2020}).
\newblock


\bibitem[\protect\citeauthoryear{Liu, Li, Wang, Zha, Su, and Huang}{Liu
  et~al\mbox{.}}{2019}]%
        {liu2019knowledge}
\bibfield{author}{\bibinfo{person}{Xuejing Liu}, \bibinfo{person}{Liang Li},
  \bibinfo{person}{Shuhui Wang}, \bibinfo{person}{Zheng-Jun Zha},
  \bibinfo{person}{Li Su}, {and} \bibinfo{person}{Qingming Huang}.}
  \bibinfo{year}{2019}\natexlab{}.
\newblock \showarticletitle{Knowledge-guided pairwise reconstruction network
  for weakly supervised referring expression grounding}. In
  \bibinfo{booktitle}{\emph{Proceedings of the 27th ACM International
  Conference on Multimedia}}. \bibinfo{pages}{539--547}.
\newblock


\bibitem[\protect\citeauthoryear{Liu, Wan, Ma, and He}{Liu
  et~al\mbox{.}}{2021}]%
        {liu2021relation}
\bibfield{author}{\bibinfo{person}{Yongfei Liu}, \bibinfo{person}{Bo Wan},
  \bibinfo{person}{Lin Ma}, {and} \bibinfo{person}{Xuming He}.}
  \bibinfo{year}{2021}\natexlab{}.
\newblock \showarticletitle{Relation-aware Instance Refinement for Weakly
  Supervised Visual Grounding}.
\newblock \bibinfo{journal}{\emph{arXiv preprint arXiv:2103.12989}}
  (\bibinfo{year}{2021}).
\newblock


\bibitem[\protect\citeauthoryear{Liu, Wan, Zhu, and He}{Liu
  et~al\mbox{.}}{2020b}]%
        {liu2020learning}
\bibfield{author}{\bibinfo{person}{Yongfei Liu}, \bibinfo{person}{Bo Wan},
  \bibinfo{person}{Xiaodan Zhu}, {and} \bibinfo{person}{Xuming He}.}
  \bibinfo{year}{2020}\natexlab{b}.
\newblock \showarticletitle{Learning cross-modal context graph for visual
  grounding}. In \bibinfo{booktitle}{\emph{Proceedings of the AAAI Conference
  on Artificial Intelligence}}, Vol.~\bibinfo{volume}{34}.
  \bibinfo{pages}{11645--11652}.
\newblock


\bibitem[\protect\citeauthoryear{Lu, Xiong, Parikh, and Socher}{Lu
  et~al\mbox{.}}{2017}]%
        {lu2017knowing}
\bibfield{author}{\bibinfo{person}{Jiasen Lu}, \bibinfo{person}{Caiming Xiong},
  \bibinfo{person}{Devi Parikh}, {and} \bibinfo{person}{Richard Socher}.}
  \bibinfo{year}{2017}\natexlab{}.
\newblock \showarticletitle{Knowing when to look: Adaptive attention via a
  visual sentinel for image captioning}. In
  \bibinfo{booktitle}{\emph{Proceedings of the IEEE conference on computer
  vision and pattern recognition}}. \bibinfo{pages}{375--383}.
\newblock


\bibitem[\protect\citeauthoryear{Ma, Kalantidis, AlRegib, Vajda, Rohrbach, and
  Kira}{Ma et~al\mbox{.}}{2020}]%
        {ma2020learning}
\bibfield{author}{\bibinfo{person}{Chih-Yao Ma}, \bibinfo{person}{Yannis
  Kalantidis}, \bibinfo{person}{Ghassan AlRegib}, \bibinfo{person}{Peter
  Vajda}, \bibinfo{person}{Marcus Rohrbach}, {and} \bibinfo{person}{Zsolt
  Kira}.} \bibinfo{year}{2020}\natexlab{}.
\newblock \showarticletitle{Learning to generate grounded visual captions
  without localization supervision}. In \bibinfo{booktitle}{\emph{Proceedings
  of the European Conference on Computer Vision (ECCV)}},
  Vol.~\bibinfo{volume}{2}. Springer.
\newblock


\bibitem[\protect\citeauthoryear{Mitchell, Dodge, Goyal, Yamaguchi, Stratos,
  Han, Mensch, Berg, Berg, and Daum{\'e}~III}{Mitchell et~al\mbox{.}}{2012}]%
        {mitchell2012midge}
\bibfield{author}{\bibinfo{person}{Margaret Mitchell}, \bibinfo{person}{Jesse
  Dodge}, \bibinfo{person}{Amit Goyal}, \bibinfo{person}{Kota Yamaguchi},
  \bibinfo{person}{Karl Stratos}, \bibinfo{person}{Xufeng Han},
  \bibinfo{person}{Alyssa Mensch}, \bibinfo{person}{Alexander Berg},
  \bibinfo{person}{Tamara Berg}, {and} \bibinfo{person}{Hal Daum{\'e}~III}.}
  \bibinfo{year}{2012}\natexlab{}.
\newblock \showarticletitle{Midge: Generating image descriptions from computer
  vision detections}. In \bibinfo{booktitle}{\emph{Proceedings of the 13th
  Conference of the European Chapter of the Association for Computational
  Linguistics}}. \bibinfo{pages}{747--756}.
\newblock


\bibitem[\protect\citeauthoryear{Pan, Gao, Lin, Tang, Dong, Yuan, Huang, and
  Xu}{Pan et~al\mbox{.}}{2021}]%
        {pan2021unveiling}
\bibfield{author}{\bibinfo{person}{Xingjia Pan}, \bibinfo{person}{Yingguo Gao},
  \bibinfo{person}{Zhiwen Lin}, \bibinfo{person}{Fan Tang},
  \bibinfo{person}{Weiming Dong}, \bibinfo{person}{Haolei Yuan},
  \bibinfo{person}{Feiyue Huang}, {and} \bibinfo{person}{Changsheng Xu}.}
  \bibinfo{year}{2021}\natexlab{}.
\newblock \showarticletitle{Unveiling the Potential of Structure-Preserving for
  Weakly Supervised Object Localization}.
\newblock \bibinfo{journal}{\emph{arXiv preprint arXiv:2103.04523}}
  (\bibinfo{year}{2021}).
\newblock


\bibitem[\protect\citeauthoryear{Papineni, Roukos, Ward, and Zhu}{Papineni
  et~al\mbox{.}}{2002}]%
        {papineni2002bleu}
\bibfield{author}{\bibinfo{person}{Kishore Papineni}, \bibinfo{person}{Salim
  Roukos}, \bibinfo{person}{Todd Ward}, {and} \bibinfo{person}{Wei-Jing Zhu}.}
  \bibinfo{year}{2002}\natexlab{}.
\newblock \showarticletitle{Bleu: a method for automatic evaluation of machine
  translation}. In \bibinfo{booktitle}{\emph{Proceedings of the 40th annual
  meeting of the Association for Computational Linguistics}}.
  \bibinfo{pages}{311--318}.
\newblock


\bibitem[\protect\citeauthoryear{Plummer, Kordas, Kiapour, Zheng, Piramuthu,
  and Lazebnik}{Plummer et~al\mbox{.}}{2018}]%
        {plummer2018conditional}
\bibfield{author}{\bibinfo{person}{Bryan~A Plummer}, \bibinfo{person}{Paige
  Kordas}, \bibinfo{person}{M~Hadi Kiapour}, \bibinfo{person}{Shuai Zheng},
  \bibinfo{person}{Robinson Piramuthu}, {and} \bibinfo{person}{Svetlana
  Lazebnik}.} \bibinfo{year}{2018}\natexlab{}.
\newblock \showarticletitle{Conditional image-text embedding networks}. In
  \bibinfo{booktitle}{\emph{Proceedings of the European Conference on Computer
  Vision (ECCV)}}. \bibinfo{pages}{249--264}.
\newblock


\bibitem[\protect\citeauthoryear{Plummer, Wang, Cervantes, Caicedo,
  Hockenmaier, and Lazebnik}{Plummer et~al\mbox{.}}{2015}]%
        {plummer2015flickr30k}
\bibfield{author}{\bibinfo{person}{Bryan~A Plummer}, \bibinfo{person}{Liwei
  Wang}, \bibinfo{person}{Chris~M Cervantes}, \bibinfo{person}{Juan~C Caicedo},
  \bibinfo{person}{Julia Hockenmaier}, {and} \bibinfo{person}{Svetlana
  Lazebnik}.} \bibinfo{year}{2015}\natexlab{}.
\newblock \showarticletitle{Flickr30k entities: Collecting region-to-phrase
  correspondences for richer image-to-sentence models}. In
  \bibinfo{booktitle}{\emph{Proceedings of the IEEE international conference on
  computer vision}}. \bibinfo{pages}{2641--2649}.
\newblock


\bibitem[\protect\citeauthoryear{Ren, He, Girshick, and Sun}{Ren
  et~al\mbox{.}}{2015}]%
        {ren2015faster}
\bibfield{author}{\bibinfo{person}{Shaoqing Ren}, \bibinfo{person}{Kaiming He},
  \bibinfo{person}{Ross Girshick}, {and} \bibinfo{person}{Jian Sun}.}
  \bibinfo{year}{2015}\natexlab{}.
\newblock \showarticletitle{Faster r-cnn: Towards real-time object detection
  with region proposal networks}.
\newblock \bibinfo{journal}{\emph{arXiv preprint arXiv:1506.01497}}
  (\bibinfo{year}{2015}).
\newblock


\bibitem[\protect\citeauthoryear{Rennie, Marcheret, Mroueh, Ross, and
  Goel}{Rennie et~al\mbox{.}}{2017}]%
        {rennie2017self}
\bibfield{author}{\bibinfo{person}{Steven~J Rennie}, \bibinfo{person}{Etienne
  Marcheret}, \bibinfo{person}{Youssef Mroueh}, \bibinfo{person}{Jerret Ross},
  {and} \bibinfo{person}{Vaibhava Goel}.} \bibinfo{year}{2017}\natexlab{}.
\newblock \showarticletitle{Self-critical sequence training for image
  captioning}. In \bibinfo{booktitle}{\emph{Proceedings of the IEEE Conference
  on Computer Vision and Pattern Recognition}}. \bibinfo{pages}{7008--7024}.
\newblock


\bibitem[\protect\citeauthoryear{Rohrbach, Rohrbach, Hu, Darrell, and
  Schiele}{Rohrbach et~al\mbox{.}}{2016}]%
        {rohrbach2016grounding}
\bibfield{author}{\bibinfo{person}{Anna Rohrbach}, \bibinfo{person}{Marcus
  Rohrbach}, \bibinfo{person}{Ronghang Hu}, \bibinfo{person}{Trevor Darrell},
  {and} \bibinfo{person}{Bernt Schiele}.} \bibinfo{year}{2016}\natexlab{}.
\newblock \showarticletitle{Grounding of textual phrases in images by
  reconstruction}. In \bibinfo{booktitle}{\emph{European Conference on Computer
  Vision}}. Springer, \bibinfo{pages}{817--834}.
\newblock


\bibitem[\protect\citeauthoryear{Shen, Ji, Yang, Deng, and Wang}{Shen
  et~al\mbox{.}}{2019}]%
        {shen2019category}
\bibfield{author}{\bibinfo{person}{Yunhang Shen}, \bibinfo{person}{Rongrong
  Ji}, \bibinfo{person}{Kuiyuan Yang}, \bibinfo{person}{Cheng Deng}, {and}
  \bibinfo{person}{Changhu Wang}.} \bibinfo{year}{2019}\natexlab{}.
\newblock \showarticletitle{Category-aware spatial constraint for weakly
  supervised detection}.
\newblock \bibinfo{journal}{\emph{IEEE Transactions on Image Processing}}
  \bibinfo{volume}{29} (\bibinfo{year}{2019}), \bibinfo{pages}{843--858}.
\newblock


\bibitem[\protect\citeauthoryear{Singh and Lee}{Singh and Lee}{2017}]%
        {singh2017hide}
\bibfield{author}{\bibinfo{person}{Krishna~Kumar Singh} {and}
  \bibinfo{person}{Yong~Jae Lee}.} \bibinfo{year}{2017}\natexlab{}.
\newblock \showarticletitle{Hide-and-seek: Forcing a network to be meticulous
  for weakly-supervised object and action localization}. In
  \bibinfo{booktitle}{\emph{IEEE ICCV}}. \bibinfo{pages}{3544--3553}.
\newblock


\bibitem[\protect\citeauthoryear{Song, Girshick, Jegelka, Mairal, Harchaoui,
  and Darrell}{Song et~al\mbox{.}}{2014}]%
        {song2014learning}
\bibfield{author}{\bibinfo{person}{Hyun~Oh Song}, \bibinfo{person}{Ross
  Girshick}, \bibinfo{person}{Stefanie Jegelka}, \bibinfo{person}{Julien
  Mairal}, \bibinfo{person}{Zaid Harchaoui}, {and} \bibinfo{person}{Trevor
  Darrell}.} \bibinfo{year}{2014}\natexlab{}.
\newblock \showarticletitle{On learning to localize objects with minimal
  supervision}. In \bibinfo{booktitle}{\emph{International Conference on
  Machine Learning}}. PMLR, \bibinfo{pages}{1611--1619}.
\newblock


\bibitem[\protect\citeauthoryear{Teh, Rochan, and Wang}{Teh
  et~al\mbox{.}}{2016}]%
        {teh2016attention}
\bibfield{author}{\bibinfo{person}{Eu~Wern Teh}, \bibinfo{person}{Mrigank
  Rochan}, {and} \bibinfo{person}{Yang Wang}.} \bibinfo{year}{2016}\natexlab{}.
\newblock \showarticletitle{Attention Networks for Weakly Supervised Object
  Localization.}. In \bibinfo{booktitle}{\emph{BMVC}}. \bibinfo{pages}{1--11}.
\newblock


\bibitem[\protect\citeauthoryear{Vaswani, Shazeer, Parmar, Uszkoreit, Jones,
  Gomez, Kaiser, and Polosukhin}{Vaswani et~al\mbox{.}}{2017}]%
        {vaswani2017attention}
\bibfield{author}{\bibinfo{person}{Ashish Vaswani}, \bibinfo{person}{Noam
  Shazeer}, \bibinfo{person}{Niki Parmar}, \bibinfo{person}{Jakob Uszkoreit},
  \bibinfo{person}{Llion Jones}, \bibinfo{person}{Aidan~N Gomez},
  \bibinfo{person}{Lukasz Kaiser}, {and} \bibinfo{person}{Illia Polosukhin}.}
  \bibinfo{year}{2017}\natexlab{}.
\newblock \showarticletitle{Attention is all you need}.
\newblock \bibinfo{journal}{\emph{arXiv preprint arXiv:1706.03762}}
  (\bibinfo{year}{2017}).
\newblock


\bibitem[\protect\citeauthoryear{Vedantam, Lawrence~Zitnick, and
  Parikh}{Vedantam et~al\mbox{.}}{2015}]%
        {vedantam2015cider}
\bibfield{author}{\bibinfo{person}{Ramakrishna Vedantam}, \bibinfo{person}{C
  Lawrence~Zitnick}, {and} \bibinfo{person}{Devi Parikh}.}
  \bibinfo{year}{2015}\natexlab{}.
\newblock \showarticletitle{Cider: Consensus-based image description
  evaluation}. In \bibinfo{booktitle}{\emph{Proceedings of the IEEE conference
  on computer vision and pattern recognition}}. \bibinfo{pages}{4566--4575}.
\newblock


\bibitem[\protect\citeauthoryear{Vinyals, Toshev, Bengio, and Erhan}{Vinyals
  et~al\mbox{.}}{2015}]%
        {vinyals2015show}
\bibfield{author}{\bibinfo{person}{Oriol Vinyals}, \bibinfo{person}{Alexander
  Toshev}, \bibinfo{person}{Samy Bengio}, {and} \bibinfo{person}{Dumitru
  Erhan}.} \bibinfo{year}{2015}\natexlab{}.
\newblock \showarticletitle{Show and tell: A neural image caption generator}.
  In \bibinfo{booktitle}{\emph{Proceedings of the IEEE conference on computer
  vision and pattern recognition}}. \bibinfo{pages}{3156--3164}.
\newblock


\bibitem[\protect\citeauthoryear{Wang, Huang, Li, Xu, Yang, and Yu}{Wang
  et~al\mbox{.}}{2020}]%
        {wang2020improving}
\bibfield{author}{\bibinfo{person}{Liwei Wang}, \bibinfo{person}{Jing Huang},
  \bibinfo{person}{Yin Li}, \bibinfo{person}{Kun Xu},
  \bibinfo{person}{Zhengyuan Yang}, {and} \bibinfo{person}{Dong Yu}.}
  \bibinfo{year}{2020}\natexlab{}.
\newblock \showarticletitle{Improving Weakly Supervised Visual Grounding by
  Contrastive Knowledge Distillation}.
\newblock \bibinfo{journal}{\emph{arXiv preprint arXiv:2007.01951}}
  (\bibinfo{year}{2020}).
\newblock


\bibitem[\protect\citeauthoryear{Wei, Shen, Cheng, Shi, Xiong, Feng, and
  Huang}{Wei et~al\mbox{.}}{2018}]%
        {wei2018ts2c}
\bibfield{author}{\bibinfo{person}{Yunchao Wei}, \bibinfo{person}{Zhiqiang
  Shen}, \bibinfo{person}{Bowen Cheng}, \bibinfo{person}{Honghui Shi},
  \bibinfo{person}{Jinjun Xiong}, \bibinfo{person}{Jiashi Feng}, {and}
  \bibinfo{person}{Thomas Huang}.} \bibinfo{year}{2018}\natexlab{}.
\newblock \showarticletitle{Ts2c: Tight box mining with surrounding
  segmentation context for weakly supervised object detection}. In
  \bibinfo{booktitle}{\emph{Proceedings of the European Conference on Computer
  Vision (ECCV)}}. \bibinfo{pages}{434--450}.
\newblock


\bibitem[\protect\citeauthoryear{Yang, Tang, Zhang, and Cai}{Yang
  et~al\mbox{.}}{2019}]%
        {yang2019auto}
\bibfield{author}{\bibinfo{person}{Xu Yang}, \bibinfo{person}{Kaihua Tang},
  \bibinfo{person}{Hanwang Zhang}, {and} \bibinfo{person}{Jianfei Cai}.}
  \bibinfo{year}{2019}\natexlab{}.
\newblock \showarticletitle{Auto-encoding scene graphs for image captioning}.
  In \bibinfo{booktitle}{\emph{Proceedings of the IEEE/CVF Conference on
  Computer Vision and Pattern Recognition}}. \bibinfo{pages}{10685--10694}.
\newblock


\bibitem[\protect\citeauthoryear{You, Jin, Wang, Fang, and Luo}{You
  et~al\mbox{.}}{2016}]%
        {you2016image}
\bibfield{author}{\bibinfo{person}{Quanzeng You}, \bibinfo{person}{Hailin Jin},
  \bibinfo{person}{Zhaowen Wang}, \bibinfo{person}{Chen Fang}, {and}
  \bibinfo{person}{Jiebo Luo}.} \bibinfo{year}{2016}\natexlab{}.
\newblock \showarticletitle{Image captioning with semantic attention}. In
  \bibinfo{booktitle}{\emph{Proceedings of the IEEE conference on computer
  vision and pattern recognition}}. \bibinfo{pages}{4651--4659}.
\newblock


\bibitem[\protect\citeauthoryear{Yu, Lin, Shen, Yang, Lu, Bansal, and Berg}{Yu
  et~al\mbox{.}}{2018}]%
        {yu2018mattnet}
\bibfield{author}{\bibinfo{person}{Licheng Yu}, \bibinfo{person}{Zhe Lin},
  \bibinfo{person}{Xiaohui Shen}, \bibinfo{person}{Jimei Yang},
  \bibinfo{person}{Xin Lu}, \bibinfo{person}{Mohit Bansal}, {and}
  \bibinfo{person}{Tamara~L Berg}.} \bibinfo{year}{2018}\natexlab{}.
\newblock \showarticletitle{Mattnet: Modular attention network for referring
  expression comprehension}. In \bibinfo{booktitle}{\emph{Proceedings of the
  IEEE Conference on Computer Vision and Pattern Recognition}}.
  \bibinfo{pages}{1307--1315}.
\newblock


\bibitem[\protect\citeauthoryear{Yu, Choi, Kim, Yoo, Lee, and Kim}{Yu
  et~al\mbox{.}}{2017}]%
        {yu2017supervising}
\bibfield{author}{\bibinfo{person}{Youngjae Yu}, \bibinfo{person}{Jongwook
  Choi}, \bibinfo{person}{Yeonhwa Kim}, \bibinfo{person}{Kyung Yoo},
  \bibinfo{person}{Sang-Hun Lee}, {and} \bibinfo{person}{Gunhee Kim}.}
  \bibinfo{year}{2017}\natexlab{}.
\newblock \showarticletitle{Supervising neural attention models for video
  captioning by human gaze data}. In \bibinfo{booktitle}{\emph{Proceedings of
  the IEEE Conference on Computer Vision and Pattern Recognition}}.
  \bibinfo{pages}{490--498}.
\newblock


\bibitem[\protect\citeauthoryear{Yun, Han, Oh, Chun, Choe, and Yoo}{Yun
  et~al\mbox{.}}{2019}]%
        {yun2019cutmix}
\bibfield{author}{\bibinfo{person}{Sangdoo Yun}, \bibinfo{person}{Dongyoon
  Han}, \bibinfo{person}{Seong~Joon Oh}, \bibinfo{person}{Sanghyuk Chun},
  \bibinfo{person}{Junsuk Choe}, {and} \bibinfo{person}{Youngjoon Yoo}.}
  \bibinfo{year}{2019}\natexlab{}.
\newblock \showarticletitle{Cutmix: Regularization strategy to train strong
  classifiers with localizable features}. In
  \bibinfo{booktitle}{\emph{Proceedings of the IEEE International Conference on
  Computer Vision}}. \bibinfo{pages}{6023--6032}.
\newblock


\bibitem[\protect\citeauthoryear{Zhang, Wang, Tang, Shi, Shi, Xiao, Zhuang, and
  Wang}{Zhang et~al\mbox{.}}{2020a}]%
        {zhang2020relational}
\bibfield{author}{\bibinfo{person}{Wenqiao Zhang}, \bibinfo{person}{Xin~Eric
  Wang}, \bibinfo{person}{Siliang Tang}, \bibinfo{person}{Haizhou Shi},
  \bibinfo{person}{Haochen Shi}, \bibinfo{person}{Jun Xiao},
  \bibinfo{person}{Yueting Zhuang}, {and} \bibinfo{person}{William~Yang Wang}.}
  \bibinfo{year}{2020}\natexlab{a}.
\newblock \showarticletitle{Relational Graph Learning for Grounded Video
  Description Generation}. In \bibinfo{booktitle}{\emph{Proceedings of the 28th
  ACM International Conference on Multimedia}}. \bibinfo{pages}{3807--3828}.
\newblock


\bibitem[\protect\citeauthoryear{Zhang, Wei, Feng, Yang, and Huang}{Zhang
  et~al\mbox{.}}{2018a}]%
        {zhang2018adversarial}
\bibfield{author}{\bibinfo{person}{Xiaolin Zhang}, \bibinfo{person}{Yunchao
  Wei}, \bibinfo{person}{Jiashi Feng}, \bibinfo{person}{Yi Yang}, {and}
  \bibinfo{person}{Thomas~S Huang}.} \bibinfo{year}{2018}\natexlab{a}.
\newblock \showarticletitle{Adversarial complementary learning for weakly
  supervised object localization}. In \bibinfo{booktitle}{\emph{IEEE CVPR}}.
  \bibinfo{pages}{1325--1334}.
\newblock


\bibitem[\protect\citeauthoryear{Zhang, Wei, Kang, Yang, and Huang}{Zhang
  et~al\mbox{.}}{2018b}]%
        {zhang2018self}
\bibfield{author}{\bibinfo{person}{Xiaolin Zhang}, \bibinfo{person}{Yunchao
  Wei}, \bibinfo{person}{Guoliang Kang}, \bibinfo{person}{Yi Yang}, {and}
  \bibinfo{person}{Thomas Huang}.} \bibinfo{year}{2018}\natexlab{b}.
\newblock \showarticletitle{Self-produced guidance for weakly-supervised object
  localization}. In \bibinfo{booktitle}{\emph{ECCV}}.
  \bibinfo{pages}{597--613}.
\newblock


\bibitem[\protect\citeauthoryear{Zhang, Wei, and Yang}{Zhang
  et~al\mbox{.}}{2020b}]%
        {zhang2020inter}
\bibfield{author}{\bibinfo{person}{Xiaolin Zhang}, \bibinfo{person}{Yunchao
  Wei}, {and} \bibinfo{person}{Yi Yang}.} \bibinfo{year}{2020}\natexlab{b}.
\newblock \showarticletitle{Inter-image communication for weakly supervised
  localization}. In \bibinfo{booktitle}{\emph{ECCV}},
  Vol.~\bibinfo{volume}{12364}. \bibinfo{pages}{271--287}.
\newblock


\bibitem[\protect\citeauthoryear{Zhou, Khosla, Lapedriza, Oliva, and
  Torralba}{Zhou et~al\mbox{.}}{2016}]%
        {zhou2016learning}
\bibfield{author}{\bibinfo{person}{Bolei Zhou}, \bibinfo{person}{Aditya
  Khosla}, \bibinfo{person}{Agata Lapedriza}, \bibinfo{person}{Aude Oliva},
  {and} \bibinfo{person}{Antonio Torralba}.} \bibinfo{year}{2016}\natexlab{}.
\newblock \showarticletitle{Learning deep features for discriminative
  localization}. In \bibinfo{booktitle}{\emph{Proceedings of the IEEE
  conference on computer vision and pattern recognition}}.
  \bibinfo{pages}{2921--2929}.
\newblock


\bibitem[\protect\citeauthoryear{Zhou, Kalantidis, Chen, Corso, and
  Rohrbach}{Zhou et~al\mbox{.}}{2019}]%
        {zhou2019grounded}
\bibfield{author}{\bibinfo{person}{Luowei Zhou}, \bibinfo{person}{Yannis
  Kalantidis}, \bibinfo{person}{Xinlei Chen}, \bibinfo{person}{Jason~J Corso},
  {and} \bibinfo{person}{Marcus Rohrbach}.} \bibinfo{year}{2019}\natexlab{}.
\newblock \showarticletitle{Grounded video description}. In
  \bibinfo{booktitle}{\emph{Proceedings of the IEEE/CVF Conference on Computer
  Vision and Pattern Recognition}}. \bibinfo{pages}{6578--6587}.
\newblock


\bibitem[\protect\citeauthoryear{Zhou, Wang, Liu, Hu, and Zhang}{Zhou
  et~al\mbox{.}}{2020}]%
        {zhou2020more}
\bibfield{author}{\bibinfo{person}{Yuanen Zhou}, \bibinfo{person}{Meng Wang},
  \bibinfo{person}{Daqing Liu}, \bibinfo{person}{Zhenzhen Hu}, {and}
  \bibinfo{person}{Hanwang Zhang}.} \bibinfo{year}{2020}\natexlab{}.
\newblock \showarticletitle{More grounded image captioning by distilling
  image-text matching model}. In \bibinfo{booktitle}{\emph{Proceedings of the
  IEEE/CVF Conference on Computer Vision and Pattern Recognition}}.
  \bibinfo{pages}{4777--4786}.
\newblock


\end{thebibliography}

\end{document}